\documentclass[10pt,journal,compsoc]{IEEEtran}

\ifCLASSOPTIONcompsoc
  \usepackage[nocompress]{cite}
\else
  \usepackage{cite}
\fi

\ifCLASSINFOpdf
\else
\fi

\usepackage{bm}
\usepackage{amsmath}
\usepackage{amssymb}
\usepackage{multirow}
\usepackage{subfigure}
\usepackage{array}
\usepackage{amsthm}
\usepackage{amsfonts}
\usepackage{graphicx}
\usepackage{rotating}

\newtheorem{definition}{Definition}
\newcommand{\citet}{\cite}
\newcommand{\figref}[1]{Figure \ref{#1}}
\newcommand{\tabref}[1]{Table \ref{#1}}
\newcommand{\secref}[1]{Section \ref{#1}}
\newcommand{\equref}[1]{Equation (\ref{#1})}

\begin{document}
%
\title{Hybrid Micro/Macro Level Convolution for Heterogeneous Graph Learning}
%
%
%

\author{Le~Yu,
        Leilei~Sun,
        Bowen~Du,
        Chuanren~Liu,
        Weifeng~Lv,
        Hui~Xiong,~\IEEEmembership{Fellow,~IEEE}
\IEEEcompsocitemizethanks{\IEEEcompsocthanksitem L. Yu, L. Sun, B. Du and W. Lv are with the SKLSDE and BDBC Lab, Beihang University, Beijing, 100083, China.\protect\\
E-mail: yule@buaa.edu.cn, leileisun@buaa.edu.cn, dubowen@buaa.edu.cn, lwf@buaa.edu.cn
\IEEEcompsocthanksitem C. Liu is with Department of Business Analytics and Statistics, University of Tennessee, Knoxville, USA.\protect\\
E-mail: cliu89@utk.edu
\IEEEcompsocthanksitem H. Xiong is with Department of Management Science and Information Systems, Rutgers University, USA.\protect\\
E-mail: hxiong@rutgers.edu}

\thanks{Manuscript received December 29, 2020; revised xx xx, xxxx.}}

%
%

\markboth{IEEE TRANSACTIONS ON KNOWLEDGE AND DATA ENGINEERING,~Vol.~XX, No.~X, XX~XXXX}%
{Yu \MakeLowercase{\textit{et al.}}: Hybrid Micro/Macro Level Convolution for Heterogeneous Graph Learning}
%

\IEEEtitleabstractindextext{%
\begin{abstract}
Heterogeneous graphs are pervasive in practical scenarios, where each graph consists of multiple types of nodes and edges.
Representation learning on heterogeneous graphs aims to obtain low-dimensional node representations that could preserve both node attributes and relation information.
However, most of the existing graph convolution approaches were designed for homogeneous graphs, and therefore cannot handle heterogeneous graphs.
Some recent methods designed for heterogeneous graphs are also faced with several issues, including the insufficient utilization of heterogeneous properties, structural information loss, and lack of interpretability.
In this paper, we propose \textit{HGConv}, a novel \textbf{H}eterogeneous \textbf{G}raph \textbf{Conv}olution approach, to learn comprehensive node representations on heterogeneous graphs with a hybrid micro/macro level convolutional operation.
Different from existing methods, HGConv could perform convolutions on the intrinsic structure of heterogeneous graphs directly at both micro and macro levels: A micro-level convolution to learn the importance of nodes within the same relation, and a macro-level convolution to distinguish the subtle difference across different relations.
The hybrid strategy enables HGConv to fully leverage heterogeneous information with proper interpretability.
Moreover, a weighted residual connection is designed to aggregate both inherent attributes and neighbor information of the focal node adaptively.
Extensive experiments on various tasks demonstrate not only the superiority of HGConv over existing methods, but also the intuitive interpretability of our approach for graph analysis.
\end{abstract}

}

\maketitle

\IEEEdisplaynontitleabstractindextext

%
\IEEEpeerreviewmaketitle

\IEEEraisesectionheading{\section{Introduction}\label{section-1}}
\IEEEPARstart{A}{} heterogeneous graph consists of multiple types of nodes and edges, involving abundant heterogeneous information \cite{DBLP:journals/sigkdd/SunH12}. In practice, heterogeneous graphs are pervasive in real-world scenarios, such as academic networks, e-commerce and social networks \cite{DBLP:journals/tkde/ShiLZSY17}. Learning meaningful representation of nodes in heterogeneous graphs is essential for various tasks, including node classification \cite{DBLP:journals/tkdd/SantosPDG18,DBLP:conf/www/ZhangXKLMZ18}, node clustering \cite{DBLP:journals/pvldb/SunAH12}, link prediction \cite{DBLP:conf/icdm/DongTWTCRC12,DBLP:conf/aaai/LiSCLTL20} and personalized recommendation \cite{DBLP:conf/wsdm/YuRSGSKNH14,DBLP:journals/tkde/ShiHZY19}.

In recent years, Graph Neural Networks (GNNs) have been widely used in representation learning of graphs and achieved superior performance. Generally, GNNs perform convolutions in two domains, namely spectral domain and spatial domain. As a spectral-based method, GCN \cite{DBLP:conf/iclr/KipfW17} utilizes the localized first-order approximation on neighbors and then performs convolutions in the Fourier domain for an entire graph. Spatial-based methods, including GraphSAGE \cite{DBLP:conf/nips/HamiltonYL17} and GAT \cite{DBLP:conf/iclr/VelickovicCCRLB18}, directly perform information propagation in the graph domain by particularly designed aggregation functions or the attention mechanism. 
However, all of the above methods were designed for homogeneous graphs with single node type and single edge type, and they are infeasible to handle the rich information in heterogeneous graphs. 
Simply adapting them to deal with heterogeneous graphs would lead to the information loss issue, since they ignore the graph heterogeneous properties.

\begin{table*}[t]
\centering
\caption{Comparison of several existing methods with the proposed model.}
\label{tab:model_analysis}
\begin{tabular}{|c|c|c|c|c|c|c|}
\hline
Models    & \begin{tabular}[c]{@{}c@{}}Graph \\ Topology\end{tabular} & \begin{tabular}[c]{@{}c@{}}Heterogeneous \\ Properties\end{tabular} & \begin{tabular}[c]{@{}c@{}}Without Specific\\ Domain Knowledge\end{tabular} & \begin{tabular}[c]{@{}c@{}}Attentive \\ Aggregation\end{tabular} & \begin{tabular}[c]{@{}c@{}}Convolutions on \\ Intrinsic Structure\end{tabular} & \begin{tabular}[c]{@{}c@{}}Multi-level \\ Representation\end{tabular} \\ \hline
MLP        & \textbf{$\times$}                                                         & \textbf{$\times$}                                                                  & \textbf{\checkmark}                                                                      & \textbf{$\times$}                                                                & \textbf{$\times$}                                                                              & \textbf{$\times$}                                                                              \\ \hline
GCN       & \textbf{\checkmark}                                                         & \textbf{$\times$}                                                                  & \textbf{\checkmark}                                                                      & \textbf{$\times$}                                                                & \textbf{\checkmark}                                                                              & \textbf{$\times$}                                                                              \\ \hline
GAT       & \textbf{\checkmark}                                                         & \textbf{$\times$}                                                                  & \textbf{\checkmark}                                                                      & \textbf{\checkmark}                                                                & \textbf{\checkmark}                                                                              & \textbf{$\times$}                                                                              \\ \hline
RGCN      & \textbf{\checkmark}                                                         & \textbf{\checkmark}                                                                  & \textbf{\checkmark}                                                                      & \textbf{$\times$}                                                                & \textbf{\checkmark}                                                                              & \textbf{$\times$}                                                                              \\ \hline
HAN       & \textbf{\checkmark}                                                         & \textbf{\checkmark}                                                                  & \textbf{$\times$}                                                                      & \textbf{\checkmark}                                                                & \textbf{\checkmark}                                                                              & \textbf{\checkmark}                                                                              \\ \hline
HetGNN    & \textbf{\checkmark}                                                         & \textbf{\checkmark}                                                                  & \textbf{\checkmark}                                                                      & \textbf{\checkmark}                                                                & \textbf{$\times$}                                                                              & \textbf{\checkmark}                                                                              \\ \hline
HGT       & \textbf{\checkmark}                                                         & \textbf{\checkmark}                                                                  & \textbf{\checkmark}                                                                      & \textbf{\checkmark}                                                                & \textbf{\checkmark}                                                                              & \textbf{$\times$}                                                                              \\ \hline
HGConv & \textbf{\checkmark}                                                         & \textbf{\checkmark}                                                                  & \textbf{\checkmark}                                                                      & \textbf{\checkmark}                                                                & \textbf{\checkmark}                                                                              & \textbf{\checkmark}  \\ \hline                                                  
\end{tabular}
\end{table*}

Despite the investigation of approaches on homogeneous graphs, there are also several attempts to design graph convolution methods for heterogeneous graphs. RGCN \cite{DBLP:conf/esws/SchlichtkrullKB18} was proposed to deal with multiple relations in knowledge graphs. HAN \cite{DBLP:conf/www/WangJSWYCY19} was designed to learn on heterogeneous graphs, which is based on meta-paths and the attention mechanism. \citet{DBLP:conf/kdd/ZhangSHSC19} presented HetGNN to consider the heterogeneity of node attributes and neighbors through dedicated aggregation functions. \citet{DBLP:conf/www/HuDWS20} proposed HGT, a variant of Transformer \cite{vaswani2017attention}, to focus on the meta relations in heterogeneous graphs. 

However, the aforementioned methods are still faced with the following limitations. 
1) \textit{Heterogeneous information loss}: several methods just utilize the properties of nodes or relations partially, rather than the comprehensive information of nodes and relations (e.g., RGCN and HAN). In detail, RGCN ignores the distinct attributes of nodes with various types. HAN relies on multiple hand-designed symmetric meta-paths to convert the heterogeneous graph into multiple homogeneous graphs, which would lead to the loss of different nodes and edges information. 
2) \textit{Structural information loss}: some methods deal with the graph topology through heuristic strategies, such as the random walk in HetGNN, which may break the intrinsic graph structure and lose valuable structural information.
3) \textit{Empirical manual design}: the performance of some methods severely relies on prior experience because of the requirement of specific domain knowledge, such as pre-defined meth-paths in HAN; 
4) \textit{Insufficient representation ability}: some methods cannot provide multi-level representation due to the flat model architecture. For example, HGT learns the interaction of nodes and relations in a single aggregation process, which is hard to distinguish their importance in such a flat architecture.

To cope with the above issues, we propose HGConv, a novel \textbf{H}eterogeneous \textbf{G}raph \textbf{Conv}olution approach, to learn node representation on heterogeneous graphs with a hybrid micro/macro level convolutional operation. 
Specifically, for a focal node: in micro-level convolution, the transformation matrices and attention vectors are both specific to node types, aiming to learn the importance of nodes within the same relation; in macro-level convolution, transformation matrices specific to relation types and the weight-sharing attention vector are employed to distinguish the subtle difference across different relations. 
Due to the hybrid micro/macro level convolution, HGConv could fully utilize the heterogeneous information of nodes and relations with proper interpretability.
Moreover, a weighted residual connection component is designed to obtain the optimal fusion of the focal node's inherent attributes and neighbor information. 
Based on the aforementioned components, our approach could be optimized in an end-to-end manner.
Comparison of several existing methods with our model are shown in \tabref{tab:model_analysis}.

To sum up, the contributions of our work are as follows:
\begin{itemize}
    \item
    \textit{A novel heterogeneous graph convolution approach} is proposed to directly perform convolutions on the intrinsic heterogeneous graph structure with a hybrid micro/macro level convolutional operation, where the micro convolution encodes the attributes of different types of nodes and the macro convolution computes on different relations respectively.

    \item
    \textit{A residual connection component with weighted combination} is designed to aggregate focal node's inherent attributes and neighbor information adaptively, which could provide comprehensive node representation.
    
    \item
    \textit{A systematic analysis on existing heterogeneous graph learning methods} is given, and we point out that each existing method could be treated as a special case of the proposed HGConv under certain circumstances.
    
\end{itemize}

The rest of this paper is organized as follows: \secref{section-2} reviews previous work related to the studied problem. \secref{section-3} introduces the studied problem. \secref{section-4} presents the framework and each component of the proposed model. \secref{section-5} evaluates the proposed model by experiments. \secref{section-6} concludes the entire paper.

\section{Related work}\label{section-2}
This section reviews existing literature related to our work and also points out their differences with our work.

\textbf{Graph Mining}.
Over the past decades, a great amount of research has been investigated on graph mining. Classical methods based on manifold learning, including Locally Linear Embedding (LLE) \cite{roweis2000nonlinear} and Laplacian Eigenmaps (LE) \cite{DBLP:conf/nips/BelkinN01}, mainly focus on the reconstruction of graphs. Inspired by the language model Skip-gram \cite{mikolov2013distributed}, more advanced methods were proposed to learn representations of nodes, such as DeepWalk \cite{perozzi2014deepwalk} and Node2Vec \cite{grover2016node2vec}. These methods adopt random walk strategy to generate sequences of nodes and use Skip-gram to maximize node co-occurrence probability in the same sequence.

However, all of the above methods only focused on the study of graph topology structure and could not take the node attributes into consideration, resulting in inferior performance. These methods are surpassed by recently proposed GNNs, which could consider both node attributes and graph structure simultaneously.

\textbf{Graph Neural Networks}.
Recent years have witnessed the success of GNNs in various tasks, such as node classification \cite{DBLP:conf/iclr/KipfW17,DBLP:conf/nips/HamiltonYL17}, link prediction \cite{DBLP:conf/nips/ZhangC18} and graph classification \cite{DBLP:conf/iclr/ErricaPBM20}. GNNs consider both graph structure and node attributes by first propagating information among each node and its neighbors, and then providing node representation based on the received information. Generally, GNNs could be divided into spectral-based methods and spatial-based methods. As a spectral-based method, Spectral CNN \cite{DBLP:journals/corr/BrunaZSL13} performs convolution in the Fourier domain by computing the eigendecomposition of the graph Laplacian matrix. ChebNet \cite{DBLP:conf/nips/DefferrardBV16} leverages the K-order Chebyshev polynomials to eliminate the need to calculate the Laplacian matrix eigenvectors.
GCN \cite{DBLP:conf/iclr/KipfW17} introduces a localized first-order approximation of ChebNet to alleviate the overfitting problem. Representative spatial-based methods include GraphSAGE \cite{DBLP:conf/nips/HamiltonYL17} and GAT \cite{DBLP:conf/iclr/VelickovicCCRLB18}. \citet{DBLP:conf/nips/HamiltonYL17} proposed GraphSAGE to propagate information in the graph domain directly and designed different functions to aggregate received information. \citet{DBLP:conf/iclr/VelickovicCCRLB18} presented GAT by introducing the attention mechanism into GNNs, which enabled GAT to select more important neighbors adaptively. 
We refer the interested readers to \cite{DBLP:journals/corr/abs-1812-08434,DBLP:journals/corr/abs-1901-00596} for more comprehensive reviews on GNNs.

However, all the above methods were designed for homogeneous graphs, and could not handle the rich information in heterogeneous graphs. In this work, we aim to propose an approach to learn on heterogeneous graphs.

\textbf{Heterogeneous Graph Neural Networks}. 
Heterogeneous graphs contain abundant information of various types of nodes and relations. Mining useful information in heterogeneous graphs is essential in practical scenarios. Recently, several graph convolution methods have been proposed for learning on heterogeneous graphs. \citet{DBLP:conf/esws/SchlichtkrullKB18} presented RGCN to learn on knowledge graphs by employing specialized transformation matrices for each type of relations. 
\citet{DBLP:conf/www/WangJSWYCY19} designed HAN by extending the attention mechanism in GAT \cite{DBLP:conf/iclr/VelickovicCCRLB18} to learn the importance of neighbors and multiple hand-designed meta-paths. 
\citet{DBLP:conf/www/0004ZMK20} considered the intermediate nodes in meta-paths, which are ignored in HAN, and proposed MAGNN to aggregate the intra-meta-path and inter-meta-path information.
HetGNN \cite{DBLP:conf/kdd/ZhangSHSC19} first samples neighbors based on random walk strategy and then uses specialized Bi-LSTMs to integrate the heterogeneous node attributes and neighbors. 
\citet{DBLP:conf/www/HuDWS20} proposed HGT to introduce type-specific transformation matrices and learn the importance of different nodes and relations based on the Transformer \cite{vaswani2017attention} architecture.

Nevertheless, there are still some limitations in the above methods, including the insufficient utilization of heterogeneous properties, structural information loss, and lack of interpretability. In this paper, we aim to cope with the issues in existing approaches and design a method to learn comprehensive node representation on heterogeneous graphs by leveraging both node attributes and relation information.

\section{Problem Formalization}\label{section-3}
This section introduces related concepts and the studied problem in this paper.
\begin{definition}
    \textbf{Heterogeneous Graph}: A heterogeneous graph is defined as a directed graph $\mathcal{G}=\left(\mathcal{V},\mathcal{E},\mathcal{A},\mathcal{R}\right)$, where $\mathcal{V}$ and $\mathcal{E}$ denote the set of nodes and edges respectively.
    Each node $v \in \mathcal{V}$ and each edge $e \in \mathcal{E}$ are associated with their type mapping functions $\phi(v):\mathcal{V} \rightarrow \mathcal{A}$ and $\varphi(e):\mathcal{E} \rightarrow \mathcal{R}$, with the constraint of $|\mathcal{A}| + |\mathcal{R}| > 2$.
\end{definition}
\begin{definition}
    \textbf{Relation}: A relation represents for the interaction schema of the source node, the target node and the connected edge. Formally, for an edge $e=(u,v)$ with source node $u$ and target node $v$, the corresponding relation $R \in \mathcal{R}$ is denoted as $\left \langle \phi(u), \varphi(e), \phi(v) \right \rangle$. The inverse of $R$ is naturally represented by $R^{-1}$, and we consider the inverse relation to propagate information of two nodes from each other. Thus, the set of edges is extended as $\mathcal{E} \cup \mathcal{E}^{-1}$ and the set of relations is extended as $\mathcal{R} \cup \mathcal{R}^{-1}$. Note that the meta-paths used in heterogeneous graph learning approaches \cite{DBLP:conf/www/WangJSWYCY19,DBLP:conf/www/0004ZMK20} are defined as sequences of relations.
\end{definition}
\begin{definition}
    \textbf{Heterogeneous Graph Representation Learning}: Given a heterogeneous graph  $\mathcal{G}=(\mathcal{V},\mathcal{E},\mathcal{A},\mathcal{R})$, where nodes with type $A \in \mathcal{A}$ are associated with the attribute matrix $\bm{X}_A \in \mathbb{R}^{|\mathcal{V}_A| \times d_A}$, the task of heterogeneous graph representation learning is to obtain the $d$-dimensional representation $\bm{h}_v \in \mathbb{R}^d$ for $v \in \mathcal{V}$, where $d \ll |V|$. The learned representations are able to capture both node attributes and relation information, which could be applied in various tasks, such as node classification, node clustering and node visualization.
\end{definition}

\section{Methodology}\label{section-4}
This section presents the framework of our proposed method and each component of the proposed method is introduced step by step.

\subsection{Framework of the Proposed Model}
\begin{figure}[!htbp]
    \centering
    \includegraphics[scale = 0.36]{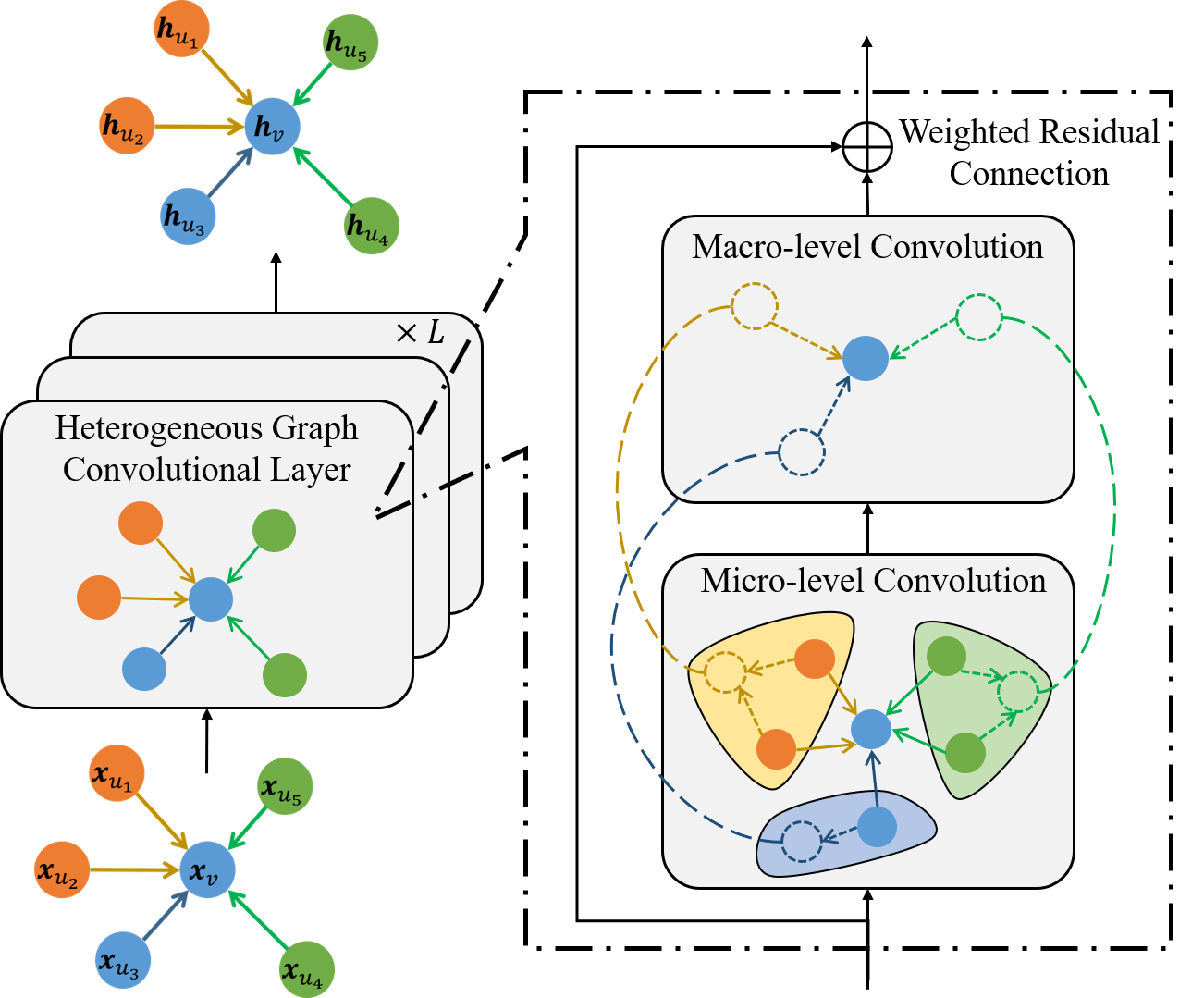}
    \caption{Framework of the proposed model.}
    \label{fig:framework}
\end{figure}

The framework of the proposed model is shown in \figref{fig:framework}, which takes the node attribute matrices $\bm{X}_A$ for $A \in \mathcal{A}$ in a heterogeneous graph as the input and provides the low-dimensional node representation $\bm{h}_v$ for $v \in \mathcal{V}$ as the output, which could be applied in various tasks.

The proposed model is made up of multiple heterogeneous graph convolutional layers, where each layer consists of the hybrid micro/macro level convolution and the weighted residual connection component. Different from \citet{DBLP:conf/www/WangJSWYCY19} that performs convolution on converted homogeneous graphs through meta-paths, the proposed hybrid convolution could directly calculate on the heterogeneous graph structure. In particular, the micro-level convolution aims to learn the importance of nodes within the same relation, and the macro-level convolution is designed to discriminate the difference across different relations. The weighted residual connection component is employed to consider the different contribution of focal node's inherent attributes and its neighbor information. By stacking multiple heterogeneous graph convolutional layers, the proposed model could consider the impacts of the focal node's directly connected and multi-hop reachable neighbors. 

\subsection{Micro-Level Convolution}
As pointed in \citet{DBLP:conf/www/WangJSWYCY19}, the importance of nodes connected with the focal node within the same relation would be different. Hence, we first design a micro-level convolution to learn the importance of nodes within the same relation. We suppose that the attributes of nodes with different types might be distributed in different latent spaces. Therefore, we utilize the transformation matrices and attention vectors, which are specific to node types, to capture the characteristics of different types of nodes in the micro-level convolution. 

Formally, we denote the focal node $v$ as the target node with type $\phi(v) \in \mathcal{A}$ and its connected node $u$ as the source node with type $\phi(u) \in \mathcal{A}$. For a focal node $v$, let $\mathcal{N}_R(v)$ denote the set of node $v$'s neighbors within $R$-type relation, where for each $u \in \mathcal{N}_R(v)$, $e = (u, v) \in \mathcal{E}$ and $R = \left \langle \phi(u), \varphi(e), \phi(v) \right \rangle \in \mathcal{R}$.

\begin{figure}[!htbp]
    \centering
    \includegraphics[scale = 0.36]{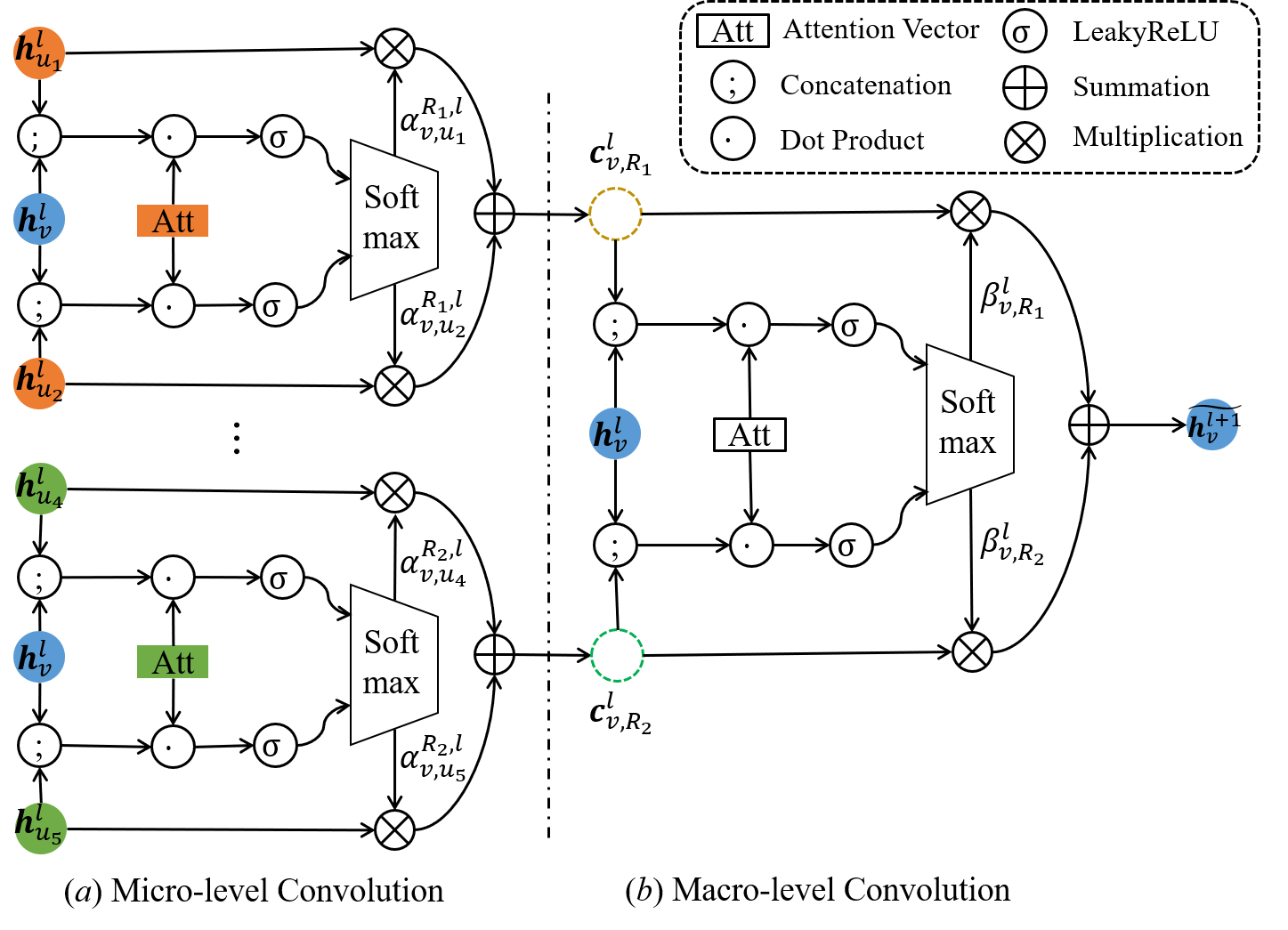}
    \caption{Explanation of the hybrid micro/macro level convolution.}
    \label{fig:hybrid_convolution}
\end{figure}

We first apply transformation matrices, which are specific to node types, to project nodes into their own latent spaces as follows,
\begin{equation}
    \bm{z}_{v}^l = \bm{W}_{\phi(v)}^l \bm{h}_{v}^{l-1},
\end{equation}
\begin{equation}
    \bm{z}_{u}^l = \bm{W}_{\phi(u)}^l \bm{h}_{u}^{l-1},
\end{equation}
where $\bm{W}_{\phi(u)}^l$ denotes the trainable transformation matrix for node $u$ with type $\phi(u)$ at layer $l$. $\bm{h}_u^l$ and $\bm{z}_u^l$ denote the original and transformed representation of node $u$ at layer $l$. Then we calculate the normalized importance of neighbor $u \in \mathcal{N}_R(v)$ as follows,
\begin{equation}
    e_{v,u}^{R,l} = LeakyReLU\left({\bm{a}_{\phi(u)}^l}^\top \left[\bm{z}_{v}^l \| \bm{z}_{u}^l \right]\right),
\end{equation}
\begin{equation}
    \alpha_{v,u}^{R,l} = \frac{\exp{\left(e_{v,u}^{R,l}\right)}}{\sum_{u\prime \in \mathcal{N}_R(v)} \exp{\left(e_{v,u\prime}^{R,l}\right)}},
\end{equation}
where $\bm{a}_{\phi(u)}^l$ is the trainable attention vector for $\phi(u)$-type source node $u$ at layer $l$ and $\|$ denotes the concatenation operation. $\top$ denotes the transpose operation. $\alpha_{v,u}^{R,l}$ is the normalized importance of source node $u$ to focal node $v$ under relation $R$ at layer $l$. Then the representation of relation $R$ about focal node $v$ is calculated by,
\begin{equation}
    \bm{c}_{v,R}^l = \sigma\left(\sum_{u \in \mathcal{N}_R(v)} \alpha_{v,u}^{R,l} \cdot \bm{z}_{u}^l\right),
\end{equation}
where $\sigma(\cdot)$ denotes the activation function (e.g., sigmoid, ReLU). 
An intuitive explanation of the micro-level convolution is shown in \figref{fig:hybrid_convolution} ($a$). Embeddings of nodes within the same relation are aggregated through the attention vectors which are specific to node types. Since the attention weight $\alpha_{v,u}^{R,l}$ is computed for each relation, it could well capture the relation information.

In order to enhance the model capacity and make the training process more stable, we employ $K$ independent heads and then concatenate representations as follows,
\begin{equation}
    \bm{c}_{v,R}^l = \mathop \| \limits_{k = 1}^K \sigma\left(\sum_{u \in \mathcal{N}_R(v)} \left[ \alpha_{v,u}^{R,l} \right]_k \cdot  \left[ \bm{z}_{u}^l \right]_k\right),
\end{equation}
where $\left[ \alpha_{v,u}^{R,l} \right]_k$ denotes the importance of source node $u$ to focal node $v$ under relation $R$ of head $k$ at layer $l$, and $\left[ \bm{z}_{u}^l \right]_k$ stands for source node $u$'s transformed representation of head $k$ at layer $l$.

\subsection{Macro-Level Convolution}
Besides considering the importance of nodes within the same relation, a focal node would also interact with multiple relations, which indicates the necessity of learning the subtle difference across different relations. Therefore, we design a macro-level convolution with the transformation matrices specific to relation types and weight-sharing attention vector to distinguish the difference of relations.

Specifically, we first transform the focal node and its connecting relations into their distinct distributed spaces by,
\begin{equation}
    {\bm{h}_{v}^l}^\prime = \bm{U}_{\phi(v)}^l \bm{h}_{v}^{l-1},
\end{equation}
\begin{equation}
    {\bm{c}_{v,R}^l}^\prime = \bm{M}_{R}^l \bm{c}_{v,R}^{l},
\end{equation}
where $\bm{U}_{\phi(v)}^l$ and $\bm{M}_{R}^l$ denote the transformation matrices for $\phi(v)$-type focal node $v$ and $R$-type relation at layer $l$ respectively. Then the normalized importance of relation $R \in \mathcal{R}(v)$ to focal node $v$ is calculated by,
\begin{equation}
    s_{v,R}^{l} = LeakyReLU\left({\bm{\mu}^{l}}^\top \left[{\bm{h}_{v}^l}^\prime \| {\bm{c}_{v,R}^l}^\prime \right]\right),
\end{equation}
\begin{equation}
    \beta_{v,R}^{l} = \frac{\exp{\left(s_{v,R}^l\right)}}{\sum_{R^{\prime} \in \mathcal{R}(v)} \exp{\left(s_{v,R^{\prime}}^l\right)}},
\end{equation}
where $\mathcal{R}(v)$ denotes the set of relations connected to focal node $v$. $\bm{\mu}^{l}$ is the trainable attention vector which is shared by different relations at layer $l$. $\beta_{v,R}^{l}$ is the normalized importance of relation $R$ to focal node $v$ at layer $l$. After obtaining the importance of different relations, we aggregate the relations as follows,
\begin{equation}
    \widetilde{\bm{h}_{v}^l} = \sum_{R \in \mathcal{R}(v)} \beta_{v,R}^{l} \cdot {\bm{c}_{v,R}^l}^\prime,
    \label{equ:aggregation_relations}
\end{equation}
where $\widetilde{\bm{h}_{v}^l}$ is the fused representation of relations connected to focal node $v$ at layer $l$. Explanation of the macro-level convolution is shown in \figref{fig:hybrid_convolution} ($b$). Representations of different relations are aggregated into a compact vector through the attention mechanism. Through the macro-level convolution, the different importance of relations could be calculated automatically.

We also extend \equref{equ:aggregation_relations} to multi-head attention by,
\begin{equation}
    \widetilde{\bm{h}_{v}^l} = \mathop \| \limits_{k = 1}^K \sum_{R \in \mathcal{R}(v)} \left[ \beta_{v,R}^{l} \right]_k \cdot \left[ {\bm{c}_{v,R}^l}^\prime \right]_k,
\end{equation}
where $\left[ \beta_{v,R}^{l} \right]_k$ is the importance of relation $R$ to focal node $v$ of head $k$ at layer $l$, and $\left[ {\bm{c}_{v,R}^l}^\prime \right]_k$ denotes the fusion of relations connected to focal node $v$ of head $k$ at layer $l$.

It is worth noting that the attention vectors in micro-level convolution are specific to node types, while in macro-level convolution, the attention vector is shared by different relations, which is unaware of relation types. Such a design is based on the following reasons. 
1) When performing micro-level convolution, nodes are associated with distinct attributes even when they are within the same relation. An attention vector unaware of node types is difficult to handle nodes' different attributes and types due to the insufficient representation ability. Hence, attention vectors specific to node types are designed in micro-level convolution. 
2) In macro-level convolution, each relation connected to the focal node is associated with a single representation and we only need to consider the difference of relation types. Therefore, the weight-sharing attention vector across different relations is designed. Following the above design, we could not only maintain the distinct characteristics of nodes and relations, but also reduce the model parameters.
    
\subsection{Weighted Residual Connection}
In addition to aggregating neighbor information by the hybrid micro/macro level convolution, the attributes of focal node are also supposed to be important, since they reflect the inherent characteristic directly. However, simply adding focal node's inherent attributes and neighbor information together could not distinguish their different importance. 

Thus, we adapt the residual connection \cite{DBLP:conf/cvpr/HeZRS16} with trainable weight parameter to combine the focal node's inherent attributes and neighbor information by,
\begin{equation}
    \bm{h}_{v}^l = \lambda_{\phi(v)}^l \cdot \bm{W}_{\phi(v),o}^l \bm{h}_{v}^{l-1} + \left(1-\lambda_{\phi(v)}^l\right) \cdot \widetilde{\bm{h}_{v}^l},
\end{equation}
where $\lambda_{\phi(v)}^l$ is the weight to control the importance of focal node $v$'s inherent attributes and its neighbor information at layer $l$. $\bm{W}_{\phi(v),o}^l$ is utilized to align the dimension of focal node $v$'s attributes and its neighbor information at layer $l$.

From another perspective, the weighted residual connection could be treated as the gated updating mechanism in Gated Recurrent Unit (GRU) \cite{DBLP:journals/corr/ChungGCB14}, where the employed update gates are specific to focal node type and carry different weights in different layers.

\subsection{The Learning Process}
We stack $L$ heterogeneous graph convolutional layers to build HGConv. For the first layer, we set $\bm{h}_v^0$ to node $v$'s corresponding row in attribute matrix $\bm{X}_{\phi(v)}$ as the input. 
The final node representation $\bm{h}_v$ is set to the output of the last layer $\bm{h}_v^L$ for $v \in \mathcal{V}$.

HGConv could be trained in an end-to-end manner with the following strategies:
1) semi-supervised learning strategy: for tasks where the labels are available, we could optimize the model by minimizing the cross entropy loss by,
\begin{equation}
    \label{equ:semi_supervised_loss}
    \mathcal{L} = - \sum_{v \in \mathcal{V}_{label}} \sum_{c=1}^{C} y_{v,c} \cdot \log \hat{y}_{v,c},
\end{equation}
where $\mathcal{V}_{label}$ is the set of nodes with labels. $y_{v,c}$ and $\hat{y}_{v,c}$ denote the ground truth and predicted possibility of node $v$ at the $c$-th dimension. In practice, $\hat{y}_{v,c}$ could be obtained from a classifier (e.g., SVM, single-layer neural network) which takes node $v$'s representation $\bm{h}_v$ as the input and outputs $\hat{\bm{y}}_v$.
2) unsupervised learning strategy: for tasks without any labels, we could optimize the model by minimizing the objective function in Skip-gram \cite{DBLP:conf/nips/MikolovSCCD13} with negative sampling,
\begin{equation}
    \label{equ:unsupervised_loss}
    \mathcal{L} = - \sum_{(v, u) \in \mathbb{S}_P} \log \sigma\left(\bm{h}_v^\top \bm{h}_u\right) - \sum_{(v^\prime, u^\prime) \in \mathbb{S}_N} \log \sigma\left(-\bm{h}_{v^\prime}^\top \bm{h}_{u^\prime}\right),
\end{equation}
where $\sigma(\cdot)$ is the sigmoid activation function, $\mathbb{S}_P$ and $\mathbb{S}_N$ denote the set of positive observed node pairs and negative sampled node pairs. 
3) joint learning strategy: we could also combine the semi-supervised and unsupervised learning strategy together to jointly optimize the model.

\subsection{Systematic Analysis of Existing Models}
Here we give a systematic analysis on existing heterogeneous graph learning models and points out that each existing method could be treated as a special case of the proposed HGConv under certain circumstances.

\textbf{Overview of Homogeneous GNNs}. 
Let us start with the introduction of homogeneous GNNs at first. Generally, the operations at the $l$-th layer of a homogeneous GNN follow a two-step strategy:
\begin{equation}
    \widetilde{\bm{h}_{v}^l} = \text{AGGREGATE}^{l}\left(\left\{\bm{h}_{u}^{l-1}: u \in \mathcal{N}\left(v\right)\right\}\right),
\end{equation}
\begin{equation}
    \bm{h}_{v}^l = \text{COMBINE}^{l}\left(\bm{h}_{v}^{l-1}, \widetilde{\bm{h}_{v}^l}\right),
    \label{equ:homogeneous_combine}
\end{equation}
where $\bm{h}_v^l$ denotes the representation of node $v$ at the $l$-th layer. $\bm{h}_v^0$ is initialized with node $v$'s original attribute $\bm{x}_v$ and $\mathcal{N}\left(v\right)$ denotes the set of node $v$'s neighbors. $\text{AGGREGATE}^l\left(\cdot\right)$ stands for the aggregation of node $v$'s neighbors. $\text{COMBINE}^l\left(\cdot\right)$ is the combination of node $v$'s inherent attribute and its neighbor information at layer $l$. 

Different architectures for AGGREGATE and COMBINE have been proposed in recent years. For example, GCN \cite{DBLP:conf/iclr/KipfW17} utilizes the normalized adjacency matrix for AGGREGATE and uses the residual connection for COMBINE. GraphSAGE \cite{DBLP:conf/nips/HamiltonYL17} designs various pooling operations for AGGREGATE and leverages the concatenation for COMBINE.

\textbf{Overview of Heterogeneous GNNs}. 
The operations in heterogeneous GNNs are based on the operations in homogeneous GNNs, with additional consideration of node attributes and relation information. Formally, the operations at the $l$-th layer could be summarized as follows:
\begin{equation}
    \bm{z}_{u}^{l} = \text{TRANSFORM}_{\phi(u)}^{l}\left(\bm{h}_{u}^{l-1}\right), \forall u \in \mathcal{V}
    \label{equ:heterogeneous_transform}
\end{equation}
\begin{equation}
    \bm{c}_{v,R}^l = \text{AGGREGATE}^{l}_{R}\left(\left\{\bm{z}_{u}^{l}: u \in \mathcal{N}_R\left(v\right)\right\}\right),
    \label{equ:heterogeneous_agg_intra_relation}
\end{equation}
\begin{equation}
    \widetilde{\bm{h}_{v}^l} = \text{AGGREGATE}^{l}\left(\left\{\bm{c}_{v,R}^l: R \in \mathcal{R}\left(v\right)\right\}\right),
    \label{equ:heterogeneous_agg_inter_relation}
\end{equation}
\begin{equation}
    \bm{h}_{v}^l = \text{COMBINE}^{l}\left(\bm{h}_{v}^{l-1}, \widetilde{\bm{h}_{v}^l}\right),
    \label{equ:heterogeneous_combine}
\end{equation}
where $\mathcal{N}_R\left(v\right)$ denotes the set of node $v$'s neighbors within $R$-type relation and $\mathcal{R}\left(v\right)$ is defined as the set of relations connected to node $v$. 

Compared with homogeneous GNNs, heterogeneous GNNs first design specialized transformation matrices for different types of nodes for TRANSFORM. Then the operations in AGGREGATE are divided into aggregation within the same relation and aggregation across different relations. Finally, the operation in COMBINE is defined as the same as \equref{equ:homogeneous_combine} in homogeneous GNNs.

\textbf{Analysis of the Proposed HGConv}.
The proposed HGConv makes delicate design for each operation in the aforementioned heterogeneous GNNs. Specifically, \equref{equ:heterogeneous_transform} - \equref{equ:heterogeneous_combine} could be rewritten as \footnote{Note that we omit the activation functions and transformation matrices for graph convolution or dimension alignment for simplicity.}:
\begin{equation}
    \bm{z}_{u}^{l} = \bm{W}_{\phi(u)}^{l}\bm{h}_{u}^{l-1}, \forall u \in \mathcal{V}
    \label{equ:hgconv_transform}
\end{equation}
\begin{equation}
    \bm{c}_{v,R}^l = \sum_{u \in \mathcal{N}_R(v)} \alpha_{v,u}^{R,l} \cdot \bm{z}_{u}^l,
    \label{equ:hgconv_agg_intra_relation}
\end{equation}
\begin{equation}
    \widetilde{\bm{h}_{v}^l} = \sum_{R \in \mathcal{R}\left(v\right)} \beta_{v,R}^{l} \cdot \bm{c}_{v,R}^l,
    \label{equ:hgconv_agg_inter_relation}
\end{equation}
\begin{equation}
    \bm{h}_{v}^l = \lambda_{\phi(v)}^l \cdot \bm{h}_{v}^{l-1} +\left (1-\lambda_{\phi(v)}^l\right) \cdot \widetilde{\bm{h}_{v}^l},
    \label{equ:hgconv_combine}
\end{equation}
where $\bm{W}_{\phi \left(u\right)}$ is the transformation matrix which is specific to node $u$'s type. $\alpha_{v,u}^{R,l}$ and $\beta_{v,R}^{l}$ are learned importance by the attention mechanism in micro-level and macro-level convolution respectively. $\lambda_{\phi(v)}^l$ is the trainable parameter to balance the importance of the focal node inherent attribute and its neighbor information.

\textbf{Connection with RGCN}.
RGCN \cite{DBLP:conf/esws/SchlichtkrullKB18} ignores distinct attributes of nodes with various types and assigns importance of neighbors within the same relation based on pre-defined constants. RGCN could be treated as a special case of the proposed HGConv with the following steps: 
1) Replace $\bm{W}_{\phi(u)}^{l}$ in \equref{equ:hgconv_transform} with identity function $I\left(\cdot\right)$, which means different distributions of node attributes with various types are not considered;
2) Replace trainable $\alpha_{v,u}^{R,l}$ in \equref{equ:hgconv_agg_intra_relation} with pre-defined constant, which is calculated by the degree of each node;
3) Set $\beta_{v,R}^{l}$ in \equref{equ:hgconv_agg_inter_relation} to $1.0$, which stands for simple sum pooling;
4) Set $\lambda_{\phi(v)}^l$ in \equref{equ:hgconv_combine} to $0.5$, which means equal contribution of node inherent attributes and neighbor information.
Note that the sum pooling operation in RGCN could not distinguish the importance of nodes and relations effectively.

\textbf{Connection with HAN}.
HAN \cite{DBLP:conf/www/WangJSWYCY19} leverages multiple symmetric meta-paths to convert the heterogeneous graph into multiple homogeneous graphs. Therefore, node $v$'s neighbors are defined by the given set of meta-paths $\Phi$. HAN could be treated as a special case of the proposed HGConv with the following steps: 
1) Replace $\bm{W}_{\phi(u)}^{l}$ in \equref{equ:hgconv_transform} with identity function $I\left(\cdot\right)$, as each converted graph only contains nodes with a single type;
2) Define the set of node $v$'s neighbors in \equref{equ:hgconv_agg_intra_relation} by meth-paths $\Phi$, that is, for each meta-path $\Phi_{i}$, the set of node $v$'s neighbors is denoted as $\mathcal{N}_{\Phi_{i}}(v)$, and then learn the importance of neighbors generated by the same meta-path through the attention mechanism;
3) Replace the aggregation of different relations in \equref{equ:hgconv_agg_inter_relation} with the aggregation of multiple meta-paths $\Phi$, and learn the importance of different meta-paths using the attention mechanism;
4) Set $\lambda_{\phi(v)}^l$ in \equref{equ:hgconv_combine} to $0.0$, which means using the neighbor information directly.
Not that the converted graphs are homogeneous, and the attributes of nodes with different types are ignored in HAN, leading to inferior performance.

\textbf{Connection with HetGNN}.
HetGNN \cite{DBLP:conf/kdd/ZhangSHSC19} leverages the random walk strategy to sample neighbors and then uses Bi-LSTMs to integrate node attributes and neighbors. Therefore, node $v$'s neighbors are generated by random walk $RW$, which could be denoted as $\mathcal{N}_{RW}(v)$. HetGNN could be treated as a special case of the proposed HGConv with the following steps: 
1) Replace $\bm{W}_{\phi(u)}^{l}$ in \equref{equ:hgconv_transform} with Bi-LSTMs to aggregate attributes of nodes with various types;
2) Define the set of node $v$'s neighbors in \equref{equ:hgconv_agg_intra_relation} by random walk $RW$ and group the neighbors by node types, that is, for each node type $t$, the set of node $v$'s neighbors is denoted as $\mathcal{N}_{RW,t}(v)$. Then, learn the importance of neighbors with the same node type through Bi-LSTMs;
3) Replace the aggregation of different relations in \equref{equ:hgconv_agg_inter_relation} with the aggregation of different node types, and learn the importance of different node types using the attention mechanism;
4) Set $\lambda_{\phi(v)}^l$ in \equref{equ:hgconv_combine} to be trainable, which is incorporated in the attention mechanism in previous step in \cite{DBLP:conf/kdd/ZhangSHSC19}.
Not that the random walk $RW$ in HetGNN may break the intrinsic graph structure and results in structural information loss.

\textbf{Connection with HGT}.
HGT \cite{DBLP:conf/www/HuDWS20} learn the importance of different nodes and relations based on the Transformer architecture by designing type-specific transformation matrices. HGT focuses on the study of each relation (a.k.a. meta relation in \cite{DBLP:conf/www/HuDWS20}), hence, the importance of source node to target node is calculated based on both the two node information as well as their connected relation in a single aggregation process. HGT could be treated as a special case of the proposed HGConv with the following steps: 
1) Replace $\bm{W}_{\phi(u)}^{l}$ in \equref{equ:hgconv_transform} with the linear projections that are specific to source node type and target node type respectively to obtain Key and Query vectors;
2) Fuse the aggregation process in \equref{equ:hgconv_agg_intra_relation} and \equref{equ:hgconv_agg_inter_relation} into a single aggregation process. The importance of source node to target node is learned from the Key and Query vectors, as well as the relation transformation matrices specific to their connected relation type;
3) Set $\lambda_{\phi(v)}^l$ in \equref{equ:hgconv_combine} to $0.5$, which means node inherent attributes and neighbor information contribute equally to the final node representation.
Not that the single aggregation process in HGT leads to a flat architecture, making it is hard to distinguish the importance of nodes and relations separately.

\section{Experiments}\label{section-5}
This section presents the experimental results on real-world datasets and detailed analysis.

\subsection{Description of Datasets}
We conduct experiments on three real-world datesets.
\begin{itemize}
	\item \textbf{ACM-3}:
    Following \citet{DBLP:conf/www/WangJSWYCY19}, we extract a subset of ACM from AMiner \footnote{https://www.aminer.cn/citation} \cite{DBLP:conf/kdd/TangZYLZS08}, which contains papers published in three areas: Data Mining (KDD, ICDM), Database (VLDB, SIGMOD) and Wireless Communication (SIGCOMM, MobiCOMM). Finally we construct a heterogeneous graph containing papers (P), authors (A) and terms (T). 
    
    \item \textbf{ACM-5}: 
    We also extract a larger subset of ACM from AMiner, which includes papers published in five areas: Data Mining (KDD, ICDM, WSDM, CIKM), Database (VLDB, ICDE), Artificial Intelligence (AAAI, IJCAI), Computer Vision (CVPR, ECCV) and Natural Language Processing (ACL, EMNLP, NAACL).
    
    \item \textbf{IMDB} \footnote{https://data.world/data-society/imdb-5000-movie-dataset}:
    We extract a subset of IMDB and consruct a heterogeneous graph containing movies (M), directors (D) and actors (A). The movies are divided into three classes: Action, Comedy, Drama.
\end{itemize}

For ACM-3 and ACM-5, we use TF-IDF \cite{ramos2003using} 
to extract keywords of the abstract and title in papers. Paper attributes are the bag-of-words representation of abstracts. Author attributes are the average representation of their published papers. Term attributes are represented as the one-hot encoding of the title keywords. For IMDB, movie attributes are the bag-of-words representation of plot keywords. Director/actor attributes are the average representation of their directing/acting movies.

Details of the datasets are summarized in \tabref{tab:dataset}.
\begin{table}[!htbp]
\centering
\scriptsize
\caption{Statistics of the datasets.}
\label{tab:dataset}
\setlength{\tabcolsep}{1.5mm}{
\begin{tabular}{|c|c|c|c|c|}
\hline
Dataset & Node                                                                                                 & Relation                                                                                  & Attribute                                                             & Data Split                                                                             \\ \hline
ACM-3   & \begin{tabular}[c]{@{}c@{}}\# Paper (P): 6,782\\ \# Author (A): 1,637\\ \# Term (T): 200\end{tabular}      & \begin{tabular}[c]{@{}c@{}}\# P-A: 13,498\\ \# P-T: 18,974\\ \# P-P: 14,925\end{tabular} & \begin{tabular}[c]{@{}c@{}}P:2,000\\ A:2,000\\ T:200\end{tabular}   & \begin{tabular}[c]{@{}c@{}}Train: 1,358\\ Validation: 678\\ Test: 4,746\end{tabular}   \\ \hline
ACM-5   & \begin{tabular}[c]{@{}c@{}}\# Paper (P): 13,328\\ \# Author (A): 2,975\\ \# Term (T): 200\end{tabular}     & \begin{tabular}[c]{@{}c@{}}\# P-A: 23,662\\ \# P-T: 36,186\\ \# P-P: 22,632\end{tabular} & \begin{tabular}[c]{@{}c@{}}P:2,000\\ A:2,000\\ T:200\end{tabular}   & \begin{tabular}[c]{@{}c@{}}Train: 2,668\\ Validation: 1,331\\ Test: 9,329\end{tabular} \\ \hline
IMDB    & \begin{tabular}[c]{@{}c@{}}\# Movie (M): 4,076\\ \# Director (D): 1,999\\ \# Actor (A): 5,069\end{tabular} & \begin{tabular}[c]{@{}c@{}}\# M-D: 4,076\\ \# M-A: 12,228\end{tabular}                  & \begin{tabular}[c]{@{}c@{}}M:1,537\\ D:1,537\\ A:1,537\end{tabular} & \begin{tabular}[c]{@{}c@{}}Train: 817\\ Validation: 407\\ Test: 2,852\end{tabular}     \\ \hline
\end{tabular}
}
\end{table}

\begin{table*}[t]
\centering
\caption{Experimental results on the node classification task.}
\label{tab:classification_results}
\begin{tabular}{|c|c|c|c|c|c|c|c|c|c|c|}
\hline
\multirow{2}{*}{Data}   & \multirow{2}{*}{Metrics}  & \multirow{2}{*}{Training} & \multirow{2}{*}{MLP} & \multirow{2}{*}{GCN} & \multirow{2}{*}{GAT} & \multirow{2}{*}{RGCN} & \multirow{2}{*}{HAN} & \multirow{2}{*}{HetGNN} & \multirow{2}{*}{HGT} & \multirow{2}{*}{HGConv} \\
                        &                           &                             &                      &                      &                      &                       &                      &                         &                      &                       \\ \hline
\multirow{10}{*}{ACM-3} & \multirow{5}{*}{Macro-F1} & 20\%                        & 0.6973               & 0.8955               & 0.8852               & 0.8981                & 0.8991               & 0.6727                  & 0.8965               & \textbf{0.9150}       \\  
                        &                           & 40\%                        & 0.7740               & 0.9012               & 0.8993               & 0.9191                & 0.9175               & 0.7736                  & 0.9188               & \textbf{0.9255}       \\  
                        &                           & 60\%                        & 0.8013               & 0.9032               & 0.9053               & 0.9262                & 0.9237               & 0.8060                  & 0.9264               & \textbf{0.9286}       \\  
                        &                           & 80\%                        & 0.8249               & 0.9068               & 0.9063               & 0.9267                & 0.9268               & 0.8242                  & \textbf{0.9329}      & 0.9306                \\  
                        &                           & 100\%                       & 0.8330               & 0.9079               & 0.9058               & 0.9299                & 0.9240               & 0.8342                  & \textbf{0.9343}      & 0.9320                \\ \cline{2-11} 
                        & \multirow{5}{*}{Micro-F1} & 20\%                        & 0.6943               & 0.8869               & 0.8754               & 0.8893                & 0.8906               & 0.6710                  & 0.8885               & \textbf{0.9089}       \\  
                        &                           & 40\%                        & 0.7710               & 0.8923               & 0.8903               & 0.9124                & 0.9103               & 0.7709                  & 0.9117               & \textbf{0.9194}       \\  
                        &                           & 60\%                        & 0.7966               & 0.8948               & 0.8968               & 0.9201                & 0.9172               & 0.8016                  & 0.9203               & \textbf{0.9221}       \\  
                        &                           & 80\%                        & 0.8205               & 0.8989               & 0.8981               & 0.9202                & 0.9205               & 0.8190                  & \textbf{0.9268}      & 0.9241                \\  
                        &                           & 100\%                       & 0.8277               & 0.9000               & 0.8979               & 0.9238                & 0.9176               & 0.8282                  & \textbf{0.9284}      & 0.9256                \\ \hline
\multirow{10}{*}{ACM-5} & \multirow{5}{*}{Macro-F1} & 20\%                        & 0.6156               & 0.8221               & 0.8253               & 0.8148                & 0.8191               & 0.6022                  & 0.8100               & \textbf{0.8270}       \\  
                        &                           & 40\%                        & 0.6585               & 0.8317               & 0.8367               & 0.8368                & 0.8404               & 0.6476                  & 0.8428               & \textbf{0.8478}       \\  
                        &                           & 60\%                        & 0.7252               & 0.8440               & 0.8441               & 0.8630                & 0.8526               & 0.7133                  & 0.8573               & \textbf{0.8701}       \\  
                        &                           & 80\%                        & 0.7503               & 0.8448               & 0.8459               & 0.8699                & 0.8610               & 0.7445                  & 0.8692               & \textbf{0.8766}       \\  
                        &                           & 100\%                       & 0.7594               & 0.8492               & 0.8466               & 0.8721                & 0.8617               & 0.7565                  & 0.8715               & \textbf{0.8792}       \\ \cline{2-11} 
                        & \multirow{5}{*}{Micro-F1} & 20\%                        & 0.6469               & 0.8364               & 0.8388               & 0.8333                & 0.8334               & 0.6420                  & 0.8286               & \textbf{0.8428}       \\  
                        &                           & 40\%                        & 0.6887               & 0.8433               & 0.8475               & 0.8501                & 0.8525               & 0.6872                  & 0.8573               & \textbf{0.8616}       \\  
                        &                           & 60\%                        & 0.7354               & 0.8545               & 0.8544               & 0.8722                & 0.8626               & 0.7248                  & 0.8668               & \textbf{0.8794}       \\  
                        &                           & 80\%                        & 0.7642               & 0.8554               & 0.8562               & 0.8809                & 0.8715               & 0.7592                  & 0.8780               & \textbf{0.8855}       \\  
                        &                           & 100\%                       & 0.7745               & 0.8597               & 0.8572               & 0.8841                & 0.8720               & 0.7721                  & 0.8825               & \textbf{0.8889}       \\ \hline
\multirow{10}{*}{IMDB}  & \multirow{5}{*}{Macro-F1} & 20\%                        & 0.4506               & 0.5003               & 0.4998               & 0.5124                & 0.5118               & 0.4281                  & 0.5171               & \textbf{0.5323}       \\  
                        &                           & 40\%                        & 0.4870               & 0.5338               & 0.5350               & 0.5578                & 0.5645               & 0.4865                  & 0.5577               & \textbf{0.5760}       \\  
                        &                           & 60\%                        & 0.5188               & 0.5559               & 0.5640               & 0.5823                & 0.5912               & 0.5110                  & 0.5781               & \textbf{0.6006}       \\  
                        &                           & 80\%                        & 0.5268               & 0.5713               & 0.5698               & 0.5939                & 0.6092               & 0.5239                  & 0.6018               & \textbf{0.6183}       \\  
                        &                           & 100\%                       & 0.5563               & 0.5845               & 0.5798               & 0.6130                & 0.6212               & 0.5453                  & 0.6159               & \textbf{0.6342}       \\ \cline{2-11} 
                        & \multirow{5}{*}{Micro-F1} & 20\%                        & 0.4598               & 0.5062               & 0.5072               & 0.5212                & 0.5263               & 0.4533                  & 0.5210               & \textbf{0.5414}       \\  
                        &                           & 40\%                        & 0.4874               & 0.5355               & 0.5378               & 0.5601                & 0.5723               & 0.4942                  & 0.5605               & \textbf{0.5792}       \\  
                        &                           & 60\%                        & 0.5186               & 0.5611               & 0.5669               & 0.5850                & 0.5968               & 0.5146                  & 0.5792               & \textbf{0.6017}       \\  
                        &                           & 80\%                        & 0.5269               & 0.5771               & 0.5757               & 0.5952                & 0.6129               & 0.5237                  & 0.6020               & \textbf{0.6193}       \\  
                        &                           & 100\%                       & 0.5538               & 0.5888               & 0.5837               & 0.6147                & 0.6242               & 0.5478                  & 0.6163               & \textbf{0.6343}       \\ \hline
\end{tabular}
\end{table*}

\begin{table*}[t]
\centering
\caption{Experimental results on the node clustering task.}
\label{tab:clustering_results}
\begin{tabular}{|c|c|c|c|c|c|c|c|c|c|c|}
\hline
Data                   & Metrics & MLP    & GCN    & GAT    & RGCN   & HAN    & HetGNN & HGT    & HGConv    & \%Improv.           \\ \hline
\multirow{2}{*}{ACM-3} & ARI     & 0.6105 & 0.7179 & 0.7319 & 0.7973 & 0.7732 & 0.6077 & 0.7944 & \textbf{0.8166} & 2.4\% \\ 
                       & NMI     & 0.5535 & 0.6806 & 0.6965 & 0.7536 & 0.7317 & 0.5520 & 0.7560 & \textbf{0.7752} & 2.5\% \\ \hline
\multirow{2}{*}{ACM-5} & ARI     & 0.5969 & 0.7010 & 0.7155 & 0.7766 & 0.7347 & 0.5931 & 0.7732 & \textbf{0.7903} & 1.8\% \\ 
                       & NMI     & 0.5501 & 0.6687 & 0.6789 & 0.7345 & 0.7056 & 0.5461 & 0.7319 & \textbf{0.7543} & 2.7\% \\ \hline
\multirow{2}{*}{IMDB}  & ARI     & 0.2011 & 0.2435 & 0.2264 & 0.3069 & 0.2777 & 0.1957 & 0.2982 & \textbf{0.3164} & 3.1\% \\ 
                       & NMI     & 0.1811 & 0.2099 & 0.2005 & 0.2647 & 0.2400 & 0.1723 & 0.2566 & \textbf{0.2757} & 4.2\% \\ \hline
\end{tabular}
\end{table*}

\subsection{Compared Methods}
We compare our method with the following baselines:
\begin{itemize}
	\item \textbf{MLP}: 
    MLP ignores the graph structure and solely focuses on the focal node attributes by leveraging the multilayer perceptron.
    
    \item \textbf{GCN}:
    GCN performs graph convolutions in the Fourier domain by leveraging the localized first-order approximation \cite{DBLP:conf/iclr/KipfW17}.
    
    \item \textbf{GAT}:
    GAT introduces the attention mechanism into GNNs and assigns different importance to the neighbors adaptively \cite{DBLP:conf/iclr/VelickovicCCRLB18}. 
    
    \item \textbf{RGCN}:
    RGCN designs specialized transformation matrices for each type of relations in the modelling of knowledge graphs \cite{DBLP:conf/esws/SchlichtkrullKB18}.
    
    \item \textbf{HAN}:
    HAN leverages the attention mechanism to aggregate neighbor information via multiple manually designed meta-paths \cite{DBLP:conf/www/WangJSWYCY19}.
    
    \item \textbf{HetGNN}:
    HetGNN considers the heterogeneity of node attributes and neighbors, and then utilizes Bi-LSTMs to integrate heterogeneous information \cite{DBLP:conf/kdd/ZhangSHSC19}.
    
    \item \textbf{HGT}:
    HGT introduces type-specific transformation matrices to capture characteristics of different nodes and relations with the Transformer architecture \cite{DBLP:conf/www/HuDWS20}.
\end{itemize}

\begin{figure*}[t]
    \centering
    \includegraphics[scale = 0.5]{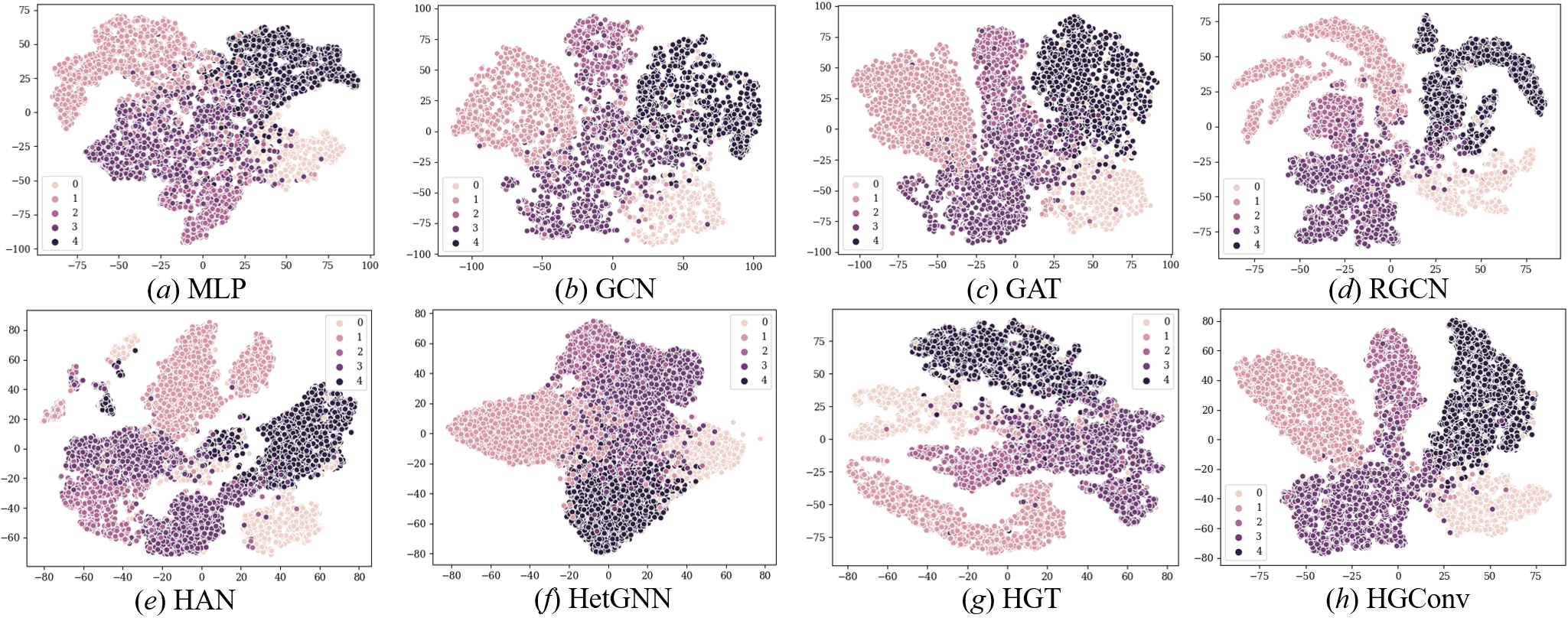}    
    \caption{Visualization of node representation on ACM-5. Each point indicates a paper and its color denotes the published area.}
    \label{fig:acm_5_node_visualization}
\end{figure*}

\subsection{Experimental Setup}
As some methods require meth-paths, we use $PAP$, $PTP$ and $PPP$ as meta-paths for ACM-3 and ACM-5, and choose $MDM$ and $MAM$ as meta-paths for IMDB. Following \citet{DBLP:conf/www/WangJSWYCY19}, we test GCN and GAT on the homogeneous graph generated by each meta-path and report the best performance from meta-paths (Experiments show that the best meta-paths on ACM-3, ACM-5, IMDB are $PAP$, $PAP$, and $MDM$ respectively). All the meta-paths are directly fed into HAN. Adam \cite{kingma2014adam} is selected as the optimizer. Dropout \cite{DBLP:journals/jmlr/SrivastavaHKSS14} is utilized to prevent over-fitting. The grid search is used to select the best hyperparameters, including dropout in $\left[0, 0.1, \cdots, 0.9\right]$ and learning rate in $\left[0.001, 0.003, 0.005, 0.008, \cdots, 0.1\right]$. The dimension of node representation is set to 64. We train all the methods with a fixed 300 epochs and use early stopping strategy with a patience of 100, which means the training process is terminated when the evaluation metrics on the validation set are not improved for 100 consecutive epochs. 

For HGConv, the numbers of attention heads in micro/macro level convolution are both set to 8, and the dimension of each head's attention vector is set to 8. We build HGConv with two layers, since two layers could achieve satisfactory performance and stacking more layers cannot improve the performance significantly. The proposed HGConv is implemented with PyTorch \footnote{https://pytorch.org/} \cite{DBLP:conf/nips/PaszkeGMLBCKLGA19} and Deep Graph Library (DGL) \footnote{https://www.dgl.ai/} \cite{wang2019deep}. Experiments are conducted on an Ubuntu machine equipped with two Intel(R) Xeon(R) CPU E5-2667 v4 @ 3.20GHz with 8 physical cores, and the GPU is NVIDIA TITAN Xp, armed with 12 GB of GDDR5X memory running at over 11 Gbps.

\subsection{Node Classification}
We conduct experiments to make comparison on the node classification task. Following \cite{DBLP:conf/www/WangJSWYCY19}, we split the datasets into training, validation and testing sets with the ratio of 2:1:7. The ratios of training data are varied in $\left[20\%, 40\%, 60\%, 80\%, 100\%\right]$. To make comprehensive comparison, we additionally use 5-fold cross-validation and report the average classification results. Macro-F1 and Micro-F1 are adopted as the evaluation metrics. For ACM-3 and ACM-5, we aim to predict the area of papers. For IMDB, the goal is to predict the class of movies. $Macro-F1$ and $Micro-F1$ are adopted as evaluation metrics. Experimental results are shown in \tabref{tab:classification_results} \footnote{Experimental results with variations and hyper-parameter settings of all the methods are shown in the appendix.}. By analyzing the results, some conclusions could be summarized.

Firstly, the performance of all the methods is improved with the increase of training data, which proves that feed more training data would help deep learning methods learn more complicated patterns and achieve better results.

Secondly, compared with MLP, the performance of other methods is significantly improved by taking graph structure into consideration in most cases, which indicates the power of graph neural networks in considering the information of both nodes and edges. 

Thirdly, methods designed for heterogeneous graphs achieve better results than methods designed for homogeneous graphs (i.e., GCN and GAT) in most cases, which demonstrates the necessity of leveraging the properties of different nodes and relations in heterogeneous graphs.

Fourthly, although HetGNN is designed for heterogeneous graph learning, it only achieves competitive or even worse results than MLP. We owe this phenomenon to the following two reasons: 1) there are several hyper-parameters (e.g., the return possibility and length of random walk, the numbers of type-grouped neighbors) in HetGNN, making the model difficult to be fine-tuned; 2) the random walk strategy may break the intrinsic graph structure and lead to structural information loss, especially when the graph structure contains valuable information.

Finally, HGConv outperforms all the baselines consistently with the varying training data ratio in most cases. Compared with MLP, GCN and GAT, HGConv takes both the graph topology and graph heterogeneity into consideration. Compared with RGCN and HAN, HGConv utilizes the specific characteristic of different nodes and relations without the requirement of domain knowledge. Compared with HetGNN, HGConv leverages intrinsic graph structure directly, which alleviates the structural information loss issue introduced by random walk. Compared with HGT, HGConv learns multi-level representation by the hybrid micro/macro level convolution, which provides HGConv with sufficient representation ability.

\subsection{Node Clustering}
The node clustering task is conducted to evaluate the learned node representations. We first obtain the node representation via feed forward on the trained model and then feed the normalized node representation into k-means algorithm. We set the number of clusters to the number of real classes for each dataset (i.e., 3, 5 and 3 for ACM-3, ACM-5 and IMDB respectively). We adopt $ARI$ and $NMI$ as evaluation metrics. Since the result of k-means tends to be affected by the initial centroids, we run k-means for 10 times and report the average results in \tabref{tab:clustering_results}.

Experimental results on the node clustering task show that HGConv outperforms all the baselines, which demonstrates the effectiveness of the learned node representation. Moreover, methods based on GNNs usually obtain better results.
We could also observe that methods achieving satisfactory results on node classification tasks (e.g., RGCN, HAN and HGT) also have satisfactory performance on node clustering tasks, which indicates that a good model could learn more universal node embedding that could be applicable to various tasks.

\subsection{Node Visualization}
To make an more intuitive comparison, we also visualize nodes in the heterogeneous graph into a low dimensional space. In particular, we project the learned node representation by HGConv into a 2-dimensional space using t-SNE \cite{maaten2008visualizing}. The visualization of node representation on ACM-5 is shown in \figref{fig:acm_5_node_visualization} \footnote{Please refer to the appendix for results on ACM-3 and IMDB.}, where the color of nodes denote their corresponding published area .

From \figref{fig:acm_5_node_visualization}, we could observe the baselines could not achieve satisfactory performance. They either fail to gather papers within the same area together, or could not provide clear boundaries of papers belonging to different areas. HGConv performs best in the visualization, as papers within the same area are closer and boundaries between different areas are more obvious.

\subsection{Ablation Study}
\begin{figure}[!htbp]
    \centering
    \includegraphics[scale = 0.36]{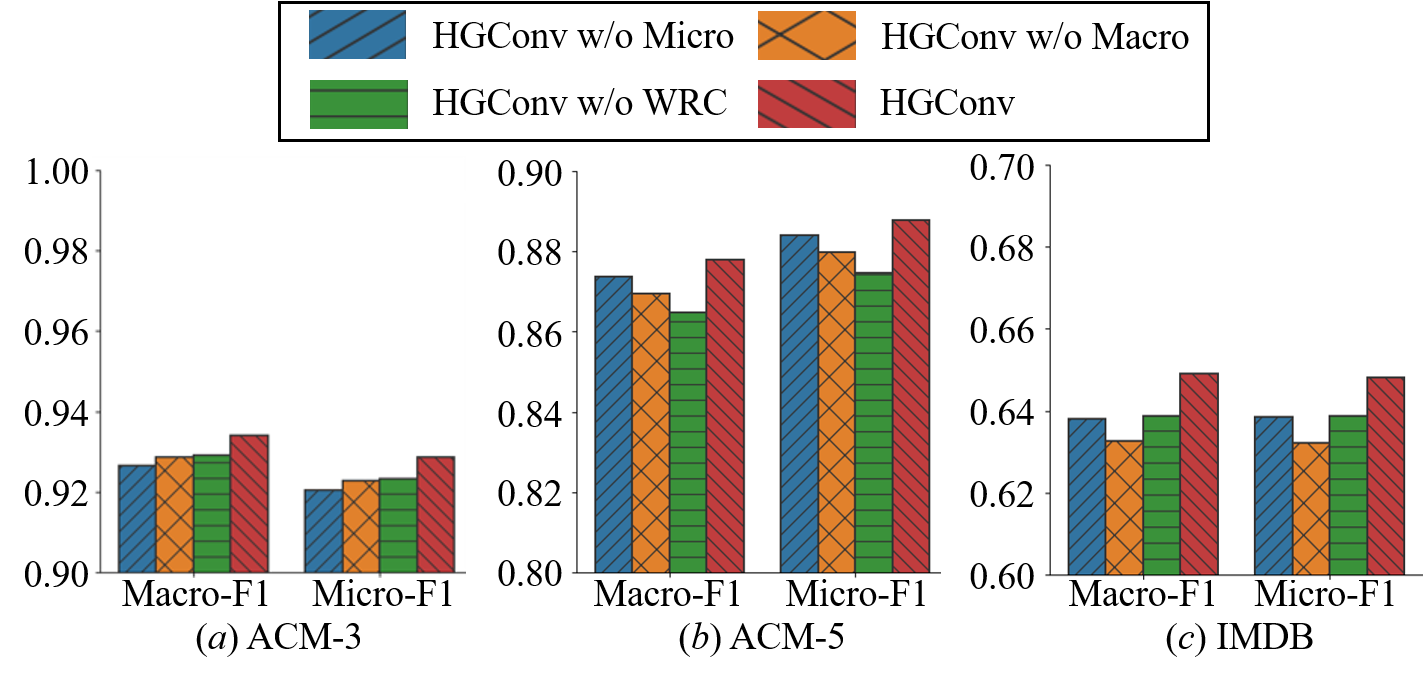}    
    \caption{Effects of the components in the proposed model.}
    \label{fig:ablation_study}
\end{figure}

We conduct the ablation study to validate the effect of each component in HGConv. We remove the micro-level convolution, macro-level convolution and weighted residual connection from HGConv respectively and denote the three variants as HGConv w/o Micro, HGConv w/o Macro and HGConv w/o WRC. 
Detailed implements of the three variants are introduced as follows:
\begin{itemize}
    \item \textbf{HGConv w/o Micro.}
    This variant replaces the micro-level convolution by performing simple average pooling on nodes within the same relation. 
    \item \textbf{HGConv w/o Macro.}
    This variant replaces the macro-level convolution by performing simple average pooling across different relations.
    \item \textbf{HGConv w/o WRC.}
    This variant removes the weighted residual connection in each layer and only uses the aggregated neighbor information as the output of each layer.
\end{itemize}
Experimental results of the variants and HGConv on the node classification task are shown in \figref{fig:ablation_study}.

\begin{figure*}[t]
    \centering
    \includegraphics[scale = 0.6]{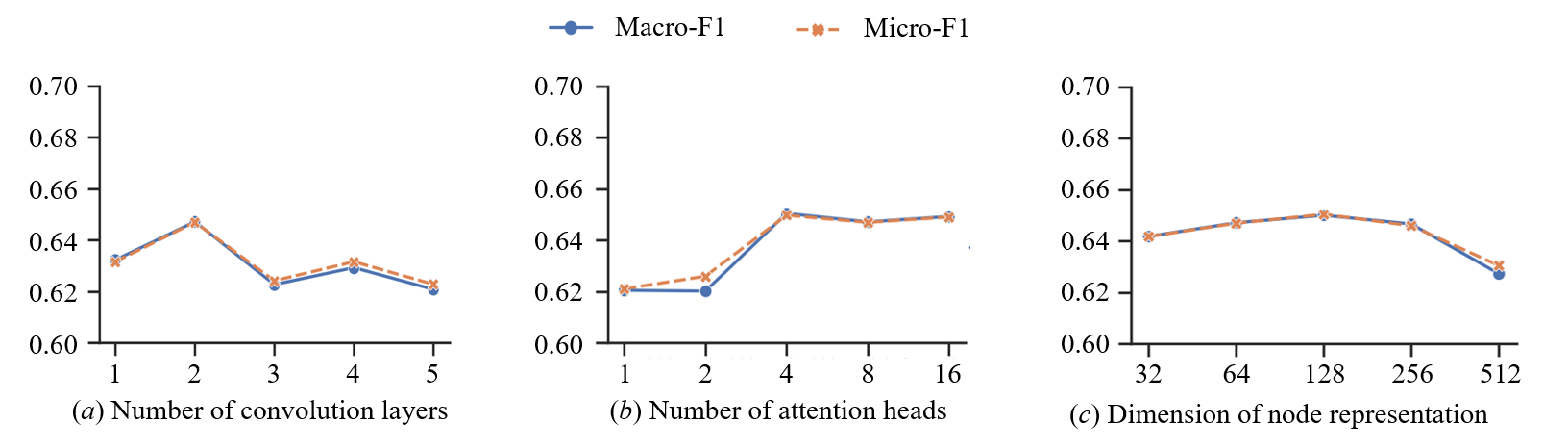}    
    \caption{Parameter Sensitivity of the proposed model on IMDB.}
    \label{fig:parameter_sensitivity}
\end{figure*}

From \figref{fig:ablation_study}, we could observe that HGConv achieves the best performance when it is equipped with all the components and removing any component would lead to worse results. The effects of the three components vary in different datasets, but all of them contribute to the improvement in the final performance. In particular, the micro-level convolution enables HGConv to select more important nodes within the same relation, and the macro-level convolution helps HGConv distinguish the subtle difference across relations. The weighted residual connection provides HGConv with the ability to consider the different contribution of focal node's inherent attributes and neighbor information. 

\subsection{Parameter Sensitivity Analysis}
We also investigate on the sensitivity analysis of several parameters in HGConv. We report the results of node classification task under different parameter settings on IMDB and experimental results are shown in \figref{fig:parameter_sensitivity}.

\textbf{Number of convolution layers}.
We build HGConv with different number of heterogeneous graph convolution layers and report the result in \figref{fig:parameter_sensitivity} ($a$). It could be observed that with the increment of layers, the performance of HGConv raises at first and then starts to drop gradually. This indicates that stacking a suitable number of layers helps the model to receive information from further neighbors, but too many layers would lead to the overfitting problem.

\textbf{Number of attention heads}.
We validate the effect of multi-head attention mechanism in the hybrid convolution by changing the number of attention heads. The result is shown in \figref{fig:parameter_sensitivity} ($b$). From the results, we could conclude that increasing the number of attention heads would improve the model performance at first. When the number of attention heads is enough (e.g., equal to or bigger than 4),  the performance reaches the top and remains stable.

\textbf{Dimension of node representation}.
We also change the dimension of node representation and report the result in \figref{fig:parameter_sensitivity} ($c$). We could find that the performance of HGConv grows with the increment of the node representation dimension and achieves the best performance when the dimension is set between 64 and 256 (we select 64 as the final setting). The performance decreases when the dimension becomes bigger further because of the overfitting problem.

\subsection{Interpretability of the Hybrid Convolution}
\begin{figure}[!htbp]
    \centering
    \includegraphics[scale = 0.45]{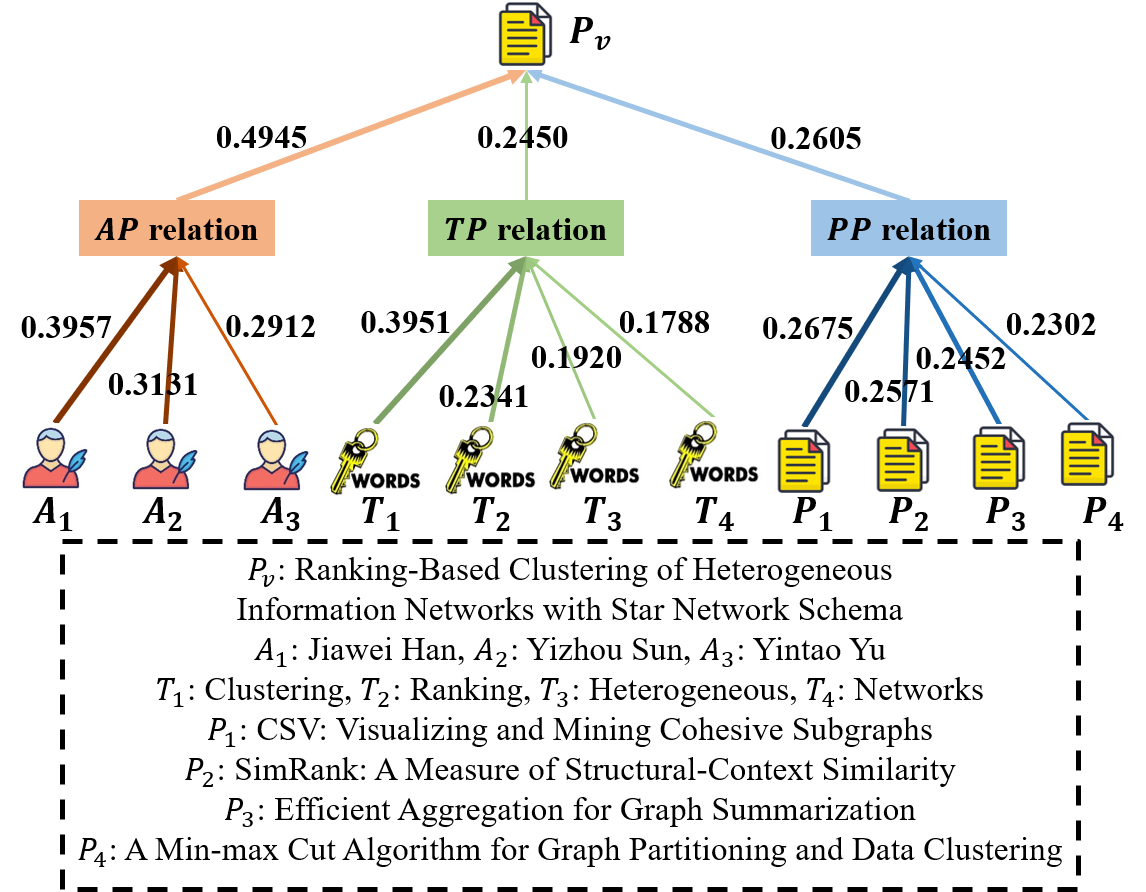}    
    \caption{Visualization of the learned attention scores.}
    \label{fig:attention_case_study}
\end{figure}

The principle components in HGConv are the micro-level convolution and macro-level convolution. Thus, we provide a detailed interpretation to better understand the learned importance of nodes within the same relation and difference across relations by the hybrid convolution. We first randomly select a sample from ACM-3 and then calculate the normalized attention scores from the last heterogeneous graph convolution layer. The selected paper $P_v$ proposes an effective ranking-based clustering algorithm for heterogeneous information network, and it is published in the Data Mining area. The visualization is shown in \figref{fig:attention_case_study}.

\textbf{Interpretation of the micro-level convolution}.  
It could be observed that in the $AP$ relation, both \textit{Jiawei Han} and \textit{Yizhou Sun} have higher attention scores than \textit{Yintao Yu} among the authors, since the first two authors contribute more in the academic research. In the $TP$ relation, keywords that are more relevant to $P_v$ (i.e., \textit{clustering} and \textit{ranking}) have higher attention scores. Moreover, the scores of references that studies more relevant topics to $P_v$ are also higher in the $PP$ relation. The above observations indicate that the micro-level convolution could select more important nodes within the same relation by assigning higher attention scores. 

\textbf{Interpretation of the macro-level convolution}.
The attention score of the $AP$ relation is much higher than that of the $TP$ or $PP$ relation, in line with the fact that GCN and GAT achieved the best performance on the $PAP$ meta-path. This finding demonstrates that the macro-level convolution could distinguish the importance of different relations automatically without empirical manual design, and the learned importance could implicitly construct more important meta-paths for specific downstream tasks.

\section{Conclusion}\label{section-6}
In this paper, we designed a hybrid micro/macro level convolution operation to address several fundamental problems in heterogeneous graph representation learning.
In particular, the micro-level convolution aims to learn the importance of nodes within the same relation and the macro-level convolution attempts to distinguish the subtle difference across different relations.
The hybrid strategy enables our model to fully leverage heterogeneous information with proper interpretability by performing convolutions on the intrinsic structure of heterogeneous graphs directly.
We also designed a weighted residual connection component to obtain the optimal combination of focal node's inherent attributes and neighbor information.
Experimental results demonstrated not only the superiority of the proposed method, but also the intuitive interpretability of our approach for graph analysis.


%

%

\ifCLASSOPTIONcompsoc
  \section*{Acknowledgments}
\else
  \section*{Acknowledgment}
\fi

This work is supported by the National Key R$\&$D Program of China [grant number 2018YFB2101003], the Science and Technology Major Project of Beijing [grant number Z191100002519012], and the National Natural Science Foundation of China [grant numbers 51778033, 51822802, 51991395, 71901011, U1811463].

\ifCLASSOPTIONcaptionsoff
  \newpage
\fi



\bibliographystyle{IEEEtran}
\bibliography{reference}

%

\begin{IEEEbiography}[{\includegraphics[width=1in,height=1.25in,clip,keepaspectratio]{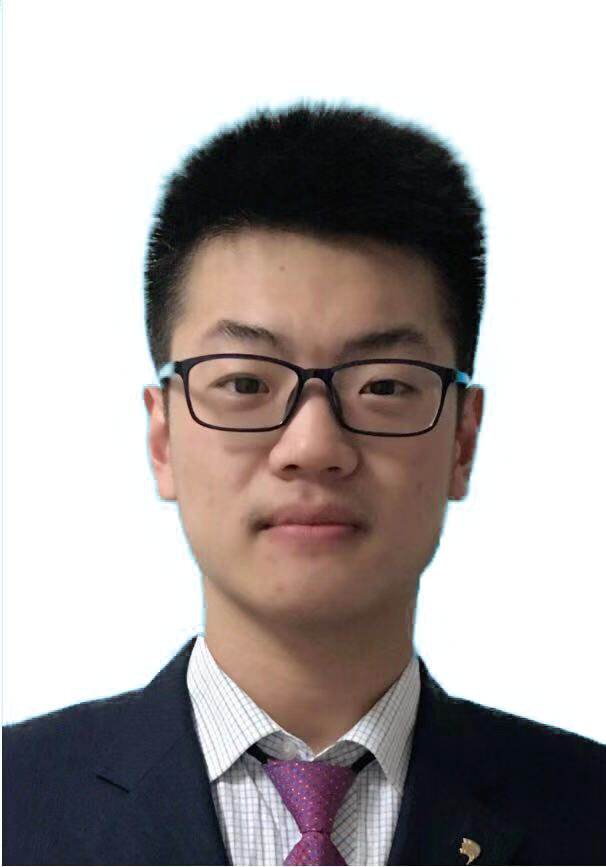}}]
{Le Yu} received the B.S. degree in Computer Science and Engineering from Beihang University, Beijing, China, in 2019. He is currently a second-year computer science Ph.D. student in the School of Computer Science and Engineering at Beihang University. His research interests include representation learning, graph neural networks and temporal data mining.
\end{IEEEbiography}

\begin{IEEEbiography}[{\includegraphics[width=1in,height=1.25in,clip,keepaspectratio]{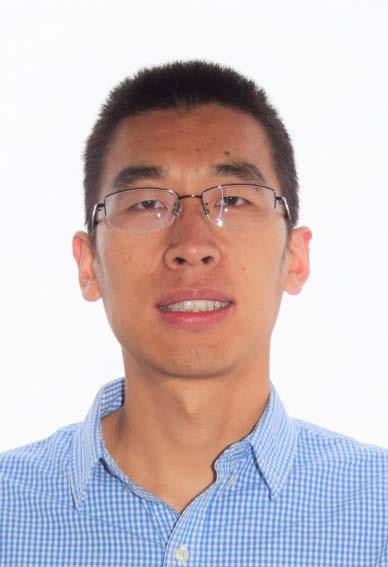}}]
{Leilei Sun} is currently an assistant professor in School of Computer Science, Beihang University, Beijing, China. He was a postdoctoral research fellow from 2017 to 2019 in School of Economics and Management, Tsinghua University. He received his Ph.D. degree from Institute of Systems Engineering, Dalian University of Technology, in 2017. His research interests include machine learning and data mining. 
\end{IEEEbiography}

\begin{IEEEbiography}[{\includegraphics[width=1in,height=1.25in,clip,keepaspectratio]{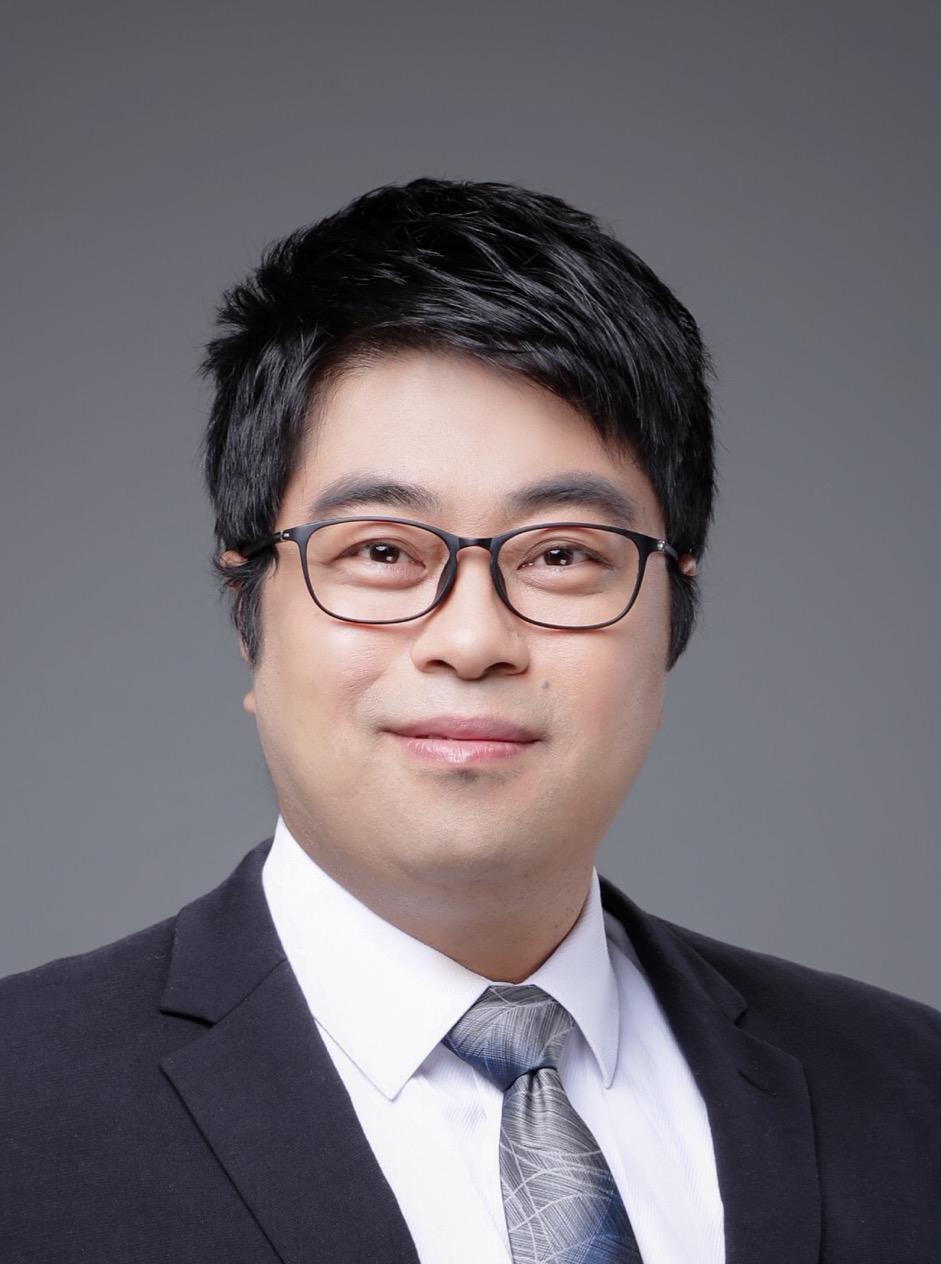}}]
{Bowen Du} received the Ph.D. degree in Computer Science and Engineering from Beihang University, Beijing, China, in 2013. He is currently a Professor with the State Key Laboratory of Software Development Environment, Beihang University. His research interests include smart city technology, multi-source data fusion, and traffic data mining.
\end{IEEEbiography}

\begin{IEEEbiography}[{\includegraphics[width=1in,height=1.25in,clip,keepaspectratio]{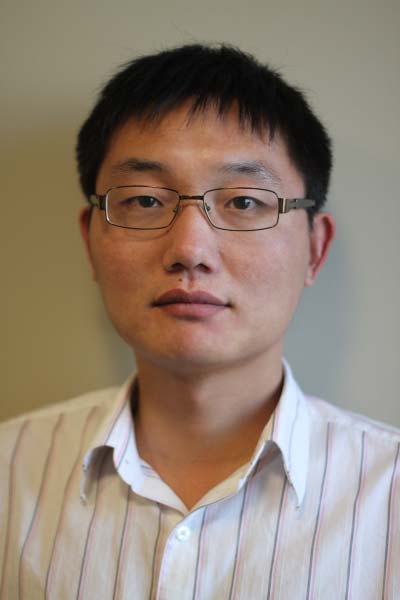}}]
{Chuanren Liu} received the B.S. degree from the University of Science and Technology of China (USTC), the M.S. degree from the Beijing University of Aeronautics and Astronautics (BUAA), and the Ph.D. degree from Rutgers, the State University of New Jersey.
He is currently an assistant professor with the Business Analytics and Statistics Department at the University of Tennessee, Knoxville, USA.
His research interests include data mining and machine learning, and their applications in business analytics.
\end{IEEEbiography}

\begin{IEEEbiography}[{\includegraphics[width=1in,height=1.25in,clip,keepaspectratio]{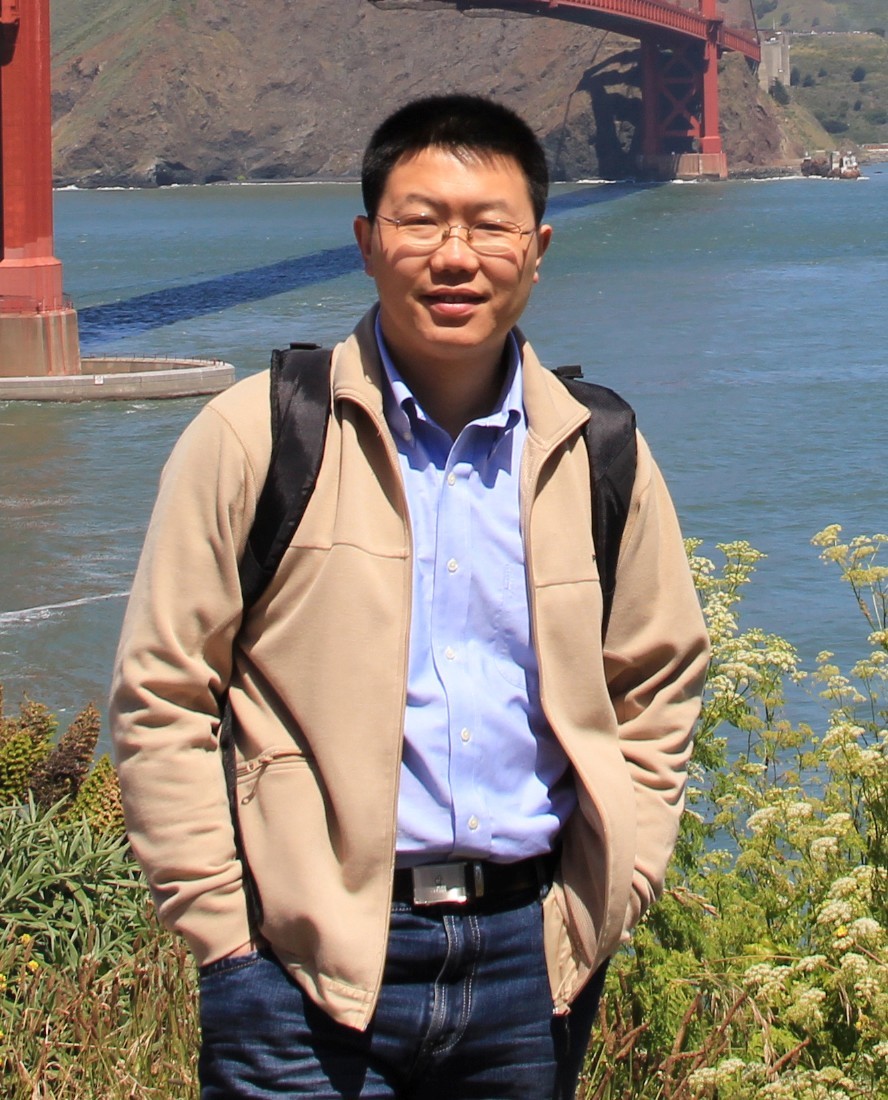}}]
{Weifeng Lv} received the B.S. degree in Computer Science and Engineering from Shandong University, Jinan, China, and the Ph.D. degree in Computer Science and Engineering from Beihang University, Beijing, China, in 1992 and 1998 respectively. Currently, he is a Professor with the State Key Laboratory of Software Development Environment, Beihang University, Beijing, China. His research interests include smart city technology and mass data processing.
\end{IEEEbiography}

\begin{IEEEbiography}[{\includegraphics[width=1in,height=1.25in,clip,keepaspectratio]{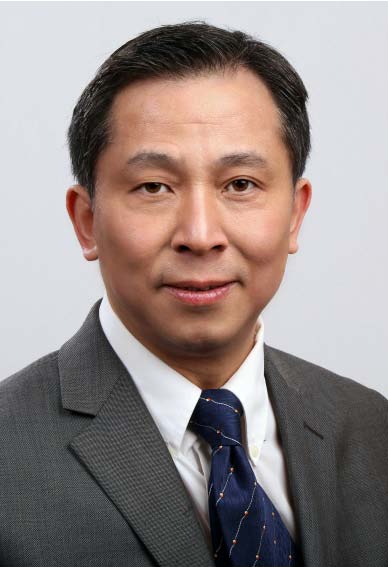}}]
{Hui Xiong} is currently a Full Professor at the Rutgers, the State University of New Jersey, where he received the 2018 Ram Charan Management Practice Award as the Grand Prix winner from the Harvard Business Review, RBS Dean’s Research Professorship (2016), the Rutgers University Board of Trustees Research Fellowship for Scholarly Excellence (2009), the ICDM Best Research Paper Award (2011), and the IEEE ICDM Outstanding Service Award (2017). He received the Ph.D. degree from the University of Minnesota (UMN), USA. He is a co-Editor-in-Chief of Encyclopedia of GIS, an Associate Editor of IEEE Transactions on Big Data (TBD), ACM Transactions on Knowledge Discovery from Data (TKDD), and ACM Transactions on Management Information Systems (TMIS). He has served regularly on the organization and program committees of numerous conferences, including as a Program Co-Chair of the Industrial and Government Track for the 18th ACM SIGKDD International Conference on Knowledge Discovery and Data Mining (KDD), a Program Co-Chair for the IEEE 2013 International Conference on Data Mining (ICDM), a General Co-Chair for the IEEE 2015 International Conference on Data Mining (ICDM), and a Program Co-Chair of the Research Track for the 2018 ACM SIGKDD International Conference on Knowledge Discovery and Data Mining. He is an IEEE Fellow and an ACM Distinguished Scientist.
\end{IEEEbiography}


\clearpage
\appendix
In the appendix, details of the experiments are introduced.

\subsection*{Node Classification}

Experimental results with variations on the node classification task are shown in \tabref{tab:classification_results_variation}.
Hyper-parameter settings are shown in \tabref{tab:hyperparameters}.

\begin{table}[!ht]
\centering
\tiny
\caption{Experimental results with variations on the node classification task.}
\label{tab:classification_results_variation}
\begin{tabular}{|c|c|c|c|c|c|c|c|c|c|c|}
\hline
\multirow{2}{*}{Data}   & \multirow{2}{*}{Metrics}  & \multirow{2}{*}{Ratio} & \multirow{2}{*}{MLP} & \multirow{2}{*}{GCN} & \multirow{2}{*}{GAT} & \multirow{2}{*}{RGCN} & \multirow{2}{*}{HAN} & \multirow{2}{*}{HetGNN} & \multirow{2}{*}{HGT}    & \multirow{2}{*}{HGConv}   \\
                        &                           &                             &                      &                      &                      &                       &                      &                         &                         &                         \\ \hline
\multirow{10}{*}{ACM-3} & \multirow{5}{*}{Macro-F1} & 20\%                        & 0.6973 $\pm$ 0.0128       & 0.8955 $\pm$ 0.0087       & 0.8852 $\pm$ 0.0131       & 0.8981 $\pm$ 0.0158        & 0.8991 $\pm$ 0.0113       & 0.6727 $\pm$ 0.0109          & 0.8965 $\pm$ 0.0083          & \textbf{0.9150 $\pm$ 0.0071} \\  
                        &                           & 40\%                        & 0.7740 $\pm$ 0.0088       & 0.9012 $\pm$ 0.0059       & 0.8993 $\pm$ 0.0062       & 0.9191 $\pm$ 0.0063        & 0.9175 $\pm$ 0.0028       & 0.7736 $\pm$ 0.0059          & 0.9188 $\pm$ 0.0052          & \textbf{0.9255 $\pm$ 0.0034} \\  
                        &                           & 60\%                        & 0.8013 $\pm$ 0.0138       & 0.9032 $\pm$ 0.0018       & 0.9053 $\pm$ 0.0039       & 0.9262 $\pm$ 0.0042        & 0.9237 $\pm$ 0.0013       & 0.8060 $\pm$ 0.0052          & 0.9264 $\pm$ 0.0025          & \textbf{0.9286 $\pm$ 0.0056} \\  
                        &                           & 80\%                        & 0.8249 $\pm$ 0.0070       & 0.9068 $\pm$ 0.0015       & 0.9063 $\pm$ 0.0065       & 0.9267 $\pm$ 0.0060        & 0.9268 $\pm$ 0.0061       & 0.8242 $\pm$  0.0037         & \textbf{0.9329 $\pm$ 0.0059} & 0.9306 $\pm$ 0.0042          \\  
                        &                           & 100\%                       & 0.8330 $\pm$ 0.0040       & 0.9079 $\pm$ 0.0010       & 0.9058 $\pm$ 0.0069       & 0.9299 $\pm$ 0.0038        & 0.9240 $\pm$ 0.0017       & 0.8342 $\pm$ 0.0043          & \textbf{0.9343 $\pm$ 0.0036} & 0.9320 $\pm$ 0.0033          \\ \cline{2-11} 
                        & \multirow{5}{*}{Micro-F1} & 20\%                        & 0.6943 $\pm$ 0.0150       & 0.8869 $\pm$ 0.0091       & 0.8754 $\pm$ 0.0141       & 0.8893 $\pm$ 0.0157        & 0.8906 $\pm$ 0.0124       & 0.6710 $\pm$ 0.0109          & 0.8885 $\pm$ 0.0084          & \textbf{0.9089 $\pm$ 0.0073} \\  
                        &                           & 40\%                        & 0.7710 $\pm$ 0.0090       & 0.8923 $\pm$ 0.0066       & 0.8903 $\pm$ 0.0068       & 0.9124 $\pm$ 0.0065        & 0.9103 $\pm$ 0.0033       & 0.7709 $\pm$ 0.0058          & 0.9117 $\pm$ 0.0058          & \textbf{0.9194 $\pm$ 0.0038} \\  
                        &                           & 60\%                        & 0.7966 $\pm$ 0.0146       & 0.8948 $\pm$ 0.0019       & 0.8968 $\pm$ 0.0041       & 0.9201 $\pm$ 0.0045        & 0.9172 $\pm$ 0.0010       & 0.8016 $\pm$ 0.0059          & 0.9203 $\pm$ 0.0032          & \textbf{0.9221 $\pm$ 0.0060} \\  
                        &                           & 80\%                        & 0.8205 $\pm$ 0.0057       & 0.8989 $\pm$ 0.0019       & 0.8981 $\pm$ 0.0068       & 0.9202 $\pm$ 0.0063        & 0.9205 $\pm$ 0.0065       & 0.8190 $\pm$ 0.0041          & \textbf{0.9268 $\pm$ 0.0063} & 0.9241 $\pm$ 0.0046          \\  
                        &                           & 100\%                       & 0.8277 $\pm$ 0.0035       & 0.9000 $\pm$ 0.0009       & 0.8979 $\pm$ 0.0074       & 0.9238 $\pm$ 0.0041        & 0.9176 $\pm$ 0.0020       & 0.8282 $\pm$ 0.0042          & \textbf{0.9284 $\pm$ 0.0037} & 0.9256 $\pm$ 0.0037          \\ \hline
\multirow{10}{*}{ACM-5} & \multirow{5}{*}{Macro-F1} & 20\%                        & 0.6156 $\pm$ 0.0127       & 0.8221 $\pm$ 0.0127       & 0.8253 $\pm$ 0.0055       & 0.8148 $\pm$ 0.0098        & 0.8191 $\pm$ 0.0019       & 0.6022 $\pm$ 0.0143          & 0.8100 $\pm$ 0.0119          & \textbf{0.8270 $\pm$ 0.0073} \\  
                        &                           & 40\%                        & 0.6585 $\pm$ 0.0321       & 0.8317 $\pm$ 0.0059       & 0.8367 $\pm$ 0.0051       & 0.8368 $\pm$ 0.0086        & 0.8404 $\pm$ 0.0072       & 0.6476 $\pm$ 0.0220          & 0.8428 $\pm$ 0.0151          & \textbf{0.8478 $\pm$ 0.0098} \\  
                        &                           & 60\%                        & 0.7252 $\pm$ 0.0097       & 0.8440 $\pm$ 0.0045       & 0.8441 $\pm$ 0.0036       & 0.8630 $\pm$ 0.0068        & 0.8526 $\pm$ 0.0046       & 0.7133 $\pm$ 0.0058          & 0.8573 $\pm$ 0.0037          & \textbf{0.8701 $\pm$ 0.0026} \\  
                        &                           & 80\%                        & 0.7503 $\pm$ 0.0055       & 0.8448 $\pm$ 0.0045       & 0.8459 $\pm$ 0.0023       & 0.8699 $\pm$ 0.0060        & 0.8610 $\pm$ 0.0029       & 0.7445 $\pm$ 0.0025          & 0.8692 $\pm$ 0.0039          & \textbf{0.8766 $\pm$ 0.0032} \\  
                        &                           & 100\%                       & 0.7594 $\pm$ 0.0055       & 0.8492 $\pm$ 0.0043       & 0.8466 $\pm$ 0.0027       & 0.8721 $\pm$ 0.0050        & 0.8617 $\pm$ 0.0030       & 0.7565 $\pm$ 0.0014          & 0.8715 $\pm$ 0.0051          & \textbf{0.8792 $\pm$ 0.0027} \\ \cline{2-11} 
                        & \multirow{5}{*}{Micro-F1} & 20\%                        & 0.6469 $\pm$ 0.0138       & 0.8364 $\pm$ 0.0112       & 0.8388 $\pm$ 0.0038       & 0.8333 $\pm$ 0.0087        & 0.8334 $\pm$ 0.0023       & 0.6420 $\pm$ 0.0076          & 0.8286 $\pm$ 0.0103          & \textbf{0.8428 $\pm$ 0.0063} \\  
                        &                           & 40\%                        & 0.6887 $\pm$ 0.0182       & 0.8433 $\pm$ 0.0054       & 0.8475 $\pm$ 0.0043       & 0.8501 $\pm$ 0.0078        & 0.8525 $\pm$ 0.0073       & 0.6872 $\pm$ 0.0101          & 0.8573 $\pm$ 0.0126          & \textbf{0.8616 $\pm$ 0.0098} \\  
                        &                           & 60\%                        & 0.7354 $\pm$ 0.0120       & 0.8545 $\pm$ 0.0043       & 0.8544 $\pm$ 0.0031       & 0.8722 $\pm$ 0.0072        & 0.8626 $\pm$ 0.0041       & 0.7248 $\pm$ 0.0051          & 0.8668 $\pm$ 0.0029          & \textbf{0.8794 $\pm$ 0.0025} \\  
                        &                           & 80\%                        & 0.7642 $\pm$ 0.0029       & 0.8554 $\pm$ 0.0048       & 0.8562 $\pm$ 0.0028       & 0.8809 $\pm$ 0.0055        & 0.8715 $\pm$ 0.0029       & 0.7592 $\pm$ 0.0029          & 0.8780 $\pm$ 0.0046          & \textbf{0.8855 $\pm$ 0.0028} \\  
                        &                           & 100\%                       & 0.7745 $\pm$ 0.0047       & 0.8597 $\pm$ 0.0037       & 0.8572 $\pm$ 0.0027       & 0.8841 $\pm$ 0.0042        & 0.8720 $\pm$ 0.0036       & 0.7721 $\pm$ 0.0008          & 0.8825 $\pm$ 0.0041          & \textbf{0.8889 $\pm$ 0.0022} \\ \hline
\multirow{10}{*}{IMDB}  & \multirow{5}{*}{Macro-F1} & 20\%                        & 0.4506 $\pm$ 0.0308       & 0.5003 $\pm$ 0.0149       & 0.4998 $\pm$ 0.0127       & 0.5124 $\pm$ 0.0184        & 0.5118 $\pm$ 0.0220       & 0.4281 $\pm$ 0.0382          & 0.5171 $\pm$ 0.0189          & \textbf{0.5323 $\pm$ 0.0255} \\  
                        &                           & 40\%                        & 0.4870 $\pm$ 0.0132       & 0.5338 $\pm$ 0.0097       & 0.5350 $\pm$ 0.0134       & 0.5578 $\pm$ 0.0127        & 0.5645 $\pm$ 0.0082       & 0.4865 $\pm$ 0.0121          & 0.5577 $\pm$ 0.0137          & \textbf{0.5760 $\pm$ 0.0076} \\  
                        &                           & 60\%                        & 0.5188 $\pm$ 0.0094       & 0.5559 $\pm$ 0.0212       & 0.5640 $\pm$ 0.0152       & 0.5823 $\pm$ 0.0135        & 0.5912 $\pm$ 0.0119       & 0.5110 $\pm$ 0.0177          & 0.5781 $\pm$ 0.0115          & \textbf{0.6006 $\pm$ 0.0107} \\  
                        &                           & 80\%                        & 0.5268 $\pm$ 0.0085       & 0.5713 $\pm$ 0.0098       & 0.5698 $\pm$ 0.0137       & 0.5939 $\pm$ 0.0080        & 0.6092 $\pm$ 0.0083       & 0.5239 $\pm$ 0.0150          & 0.6018 $\pm$ 0.0074          & \textbf{0.6183 $\pm$ 0.0054} \\  
                        &                           & 100\%                       & 0.5563 $\pm$ 0.0080       & 0.5845 $\pm$ 0.0098       & 0.5798 $\pm$ 0.0164       & 0.6130 $\pm$ 0.0085        & 0.6212 $\pm$ 0.0118       & 0.5453 $\pm$ 0.0079          & 0.6159 $\pm$ 0.0070          & \textbf{0.6342 $\pm$ 0.0115} \\ \cline{2-11} 
                        & \multirow{5}{*}{Micro-F1} & 20\%                        & 0.4598 $\pm$ 0.0171       & 0.5062 $\pm$ 0.0157       & 0.5072 $\pm$ 0.0145       & 0.5212 $\pm$ 0.0134        & 0.5263 $\pm$ 0.0122       & 0.4533 $\pm$ 0.0125          & 0.5210 $\pm$ 0.0198          & \textbf{0.5414 $\pm$ 0.0163} \\  
                        &                           & 40\%                        & 0.4874 $\pm$ 0.0129       & 0.5355 $\pm$ 0.0103       & 0.5378 $\pm$ 0.0122       & 0.5601 $\pm$ 0.0119        & 0.5723 $\pm$ 0.0088       & 0.4942 $\pm$ 0.0094          & 0.5605 $\pm$ 0.0136          & \textbf{0.5792 $\pm$ 0.0055} \\  
                        &                           & 60\%                        & 0.5186 $\pm$ 0.0095       & 0.5611 $\pm$ 0.0176       & 0.5669 $\pm$ 0.0135       & 0.5850 $\pm$ 0.0114        & 0.5968 $\pm$ 0.0090       & 0.5146 $\pm$ 0.0168          & 0.5792 $\pm$ 0.0134          & \textbf{0.6017 $\pm$ 0.0095} \\  
                        &                           & 80\%                        & 0.5269 $\pm$ 0.0101       & 0.5771 $\pm$ 0.0070       & 0.5757 $\pm$ 0.0122       & 0.5952 $\pm$ 0.0076        & 0.6129 $\pm$ 0.0075       & 0.5237 $\pm$ 0.0143          & 0.6020 $\pm$ 0.0072          & \textbf{0.6193 $\pm$ 0.0045} \\  
                        &                           & 100\%                       & 0.5538 $\pm$ 0.0083       & 0.5888 $\pm$ 0.0087       & 0.5837 $\pm$ 0.0142       & 0.6147 $\pm$ 0.0071        & 0.6242 $\pm$ 0.0125       & 0.5478 $\pm$ 0.0051          & 0.6163 $\pm$ 0.0073          & \textbf{0.6343 $\pm$ 0.0107} \\ \hline
\end{tabular}
\end{table}

\begin{table}[!ht]
\centering
\caption{Hyper-parameter settings of all the methods.}
\label{tab:hyperparameters}
\begin{tabular}{|c|c|c|c|c|c|c|c|c|c|c|}
\hline
\multirow{2}{*}{Data}   & \multirow{2}{*}{Hyper-parameters} & \multirow{2}{*}{Data ratio} & \multirow{2}{*}{MLP} & \multirow{2}{*}{GCN} & \multirow{2}{*}{GAT} & \multirow{2}{*}{RGCN} & \multirow{2}{*}{HAN} & \multirow{2}{*}{HetGNN} & \multirow{2}{*}{HGT} & \multirow{2}{*}{HGConv} \\
                        &                                   &                             &                      &                      &                      &                       &                      &                         &                      &                       \\ \hline
\multirow{10}{*}{ACM-3} & \multirow{5}{*}{learning rate}    & 20\%                        & 0.05                 & 0.05                 & 0.005                & 0.03                  & 0.01                 & 0.03                    & 0.01                 & 0.008                 \\  
                        &                                   & 40\%                        & 0.03                 & 0.005                & 0.05                 & 0.003                 & 0.01                 & 0.01                    & 0.008                & 0.005                 \\  
                        &                                   & 60\%                        & 0.03                 & 0.01                 & 0.03                 & 0.005                 & 0.01                 & 0.01                    & 0.008                & 0.003                 \\  
                        &                                   & 80\%                        & 0.05                 & 0.01                 & 0.05                 & 0.03                  & 0.01                 & 0.003                   & 0.008                & 0.001                 \\  
                        &                                   & 100\%                       & 0.005                & 0.01                 & 0.05                 & 0.03                  & 0.08                 & 0.003                   & 0.003                & 0.005                 \\ \cline{2-11} 
                        & \multirow{5}{*}{dropout}          & 20\%                        & 0.5                  & 0.1                  & 0.7                  & 0.5                   & 0.7                  & 0.5                     & 0.7                  & 0.7                   \\  
                        &                                   & 40\%                        & 0.9                  & 0.0                    & 0.6                  & 0.7                   & 0.8                  & 0.9                     & 0.8                  & 0.8                   \\  
                        &                                   & 60\%                        & 0.9                  & 0.0                    & 0.7                  & 0.7                   & 0.8                  & 0.9                     & 0.7                  & 0.6                   \\  
                        &                                   & 80\%                        & 0.9                  & 0.2                  & 0.7                  & 0.5                   & 0.7                  & 0.9                     & 0.7                  & 0.6                   \\  
                        &                                   & 100\%                       & 0.9                  & 0.5                  & 0.8                  & 0.5                   & 0.6                  & 0.9                     & 0.9                  & 0.8                   \\ \hline
\multirow{10}{*}{ACM-5} & \multirow{5}{*}{learning rate}    & 20\%                        & 0.01                 & 0.005                & 0.01                 & 0.03                  & 0.01                 & 0.01                    & 0.008                & 0.005                 \\  
                        &                                   & 40\%                        & 0.03                 & 0.01                 & 0.05                 & 0.005                 & 0.05                 & 0.01                    & 0.01                 & 0.005                 \\  
                        &                                   & 60\%                        & 0.008                & 0.03                 & 0.03                 & 0.003                 & 0.08                 & 0.01                    & 0.01                 & 0.003                 \\  
                        &                                   & 80\%                        & 0.01                 & 0.01                 & 0.005                & 0.003                 & 0.05                 & 0.01                    & 0.01                 & 0.003                 \\  
                        &                                   & 100\%                       & 0.008                & 0.005                & 0.03                 & 0.001                 & 0.01                 & 0.01                    & 0.01                 & 0.008                 \\ \cline{2-11} 
                        & \multirow{5}{*}{dropout}          & 20\%                        & 0.8                  & 0.5                  & 0.6                  & 0.5                   & 0.5                  & 0.8                     & 0.8                  & 0.5                   \\  
                        &                                   & 40\%                        & 0.8                  & 0.5                  & 0.5                  & 0.5                   & 0.5                  & 0.8                     & 0.9                  & 0.7                   \\  
                        &                                   & 60\%                        & 0.9                  & 0.2                  & 0.7                  & 0.6                   & 0.8                  & 0.8                     & 0.6                  & 0.8                   \\  
                        &                                   & 80\%                        & 0.8                  & 0.4                  & 0.5                  & 0.6                   & 0.9                  & 0.8                     & 0.7                  & 0.8                   \\  
                        &                                   & 100\%                       & 0.9                  & 0.0                    & 0.6                  & 0.5                   & 0.9                  & 0.8                     & 0.6                  & 0.8                   \\ \hline
\multirow{10}{*}{IMDB}  & \multirow{5}{*}{learning rate}    & 20\%                        & 0.01                 & 0.01                 & 0.03                 & 0.01                  & 0.08                 & 0.01                    & 0.01                 & 0.001                 \\  
                        &                                   & 40\%                        & 0.05                 & 0.08                 & 0.01                 & 0.005                 & 0.01                 & 0.01                    & 0.008                & 0.008                 \\  
                        &                                   & 60\%                        & 0.01                 & 0.003                & 0.001                & 0.03                  & 0.001                & 0.01                    & 0.001                & 0.001                 \\  
                        &                                   & 80\%                        & 0.001                & 0.05                 & 0.001                & 0.01                  & 0.05                 & 0.01                    & 0.01                 & 0.005                 \\  
                        &                                   & 100\%                       & 0.03                 & 0.05                 & 0.003                & 0.005                 & 0.05                 & 0.003                   & 0.001                & 0.003                 \\ \cline{2-11} 
                        & \multirow{5}{*}{dropout}          & 20\%                        & 0.5                  & 0.1                  & 0.5                  & 0.6                   & 0.7                  & 0.5                     & 0.2                  & 0.4                   \\  
                        &                                   & 40\%                        & 0.9                  & 0.4                  & 0.7                  & 0.5                   & 0.7                  & 0.4                     & 0.6                  & 0.5                   \\  
                        &                                   & 60\%                        & 0.9                  & 0.2                  & 0.7                  & 0.5                   & 0.7                  & 0.4                     & 0.5                  & 0.5                   \\  
                        &                                   & 80\%                        & 0.7                  & 0.3                  & 0.7                  & 0.6                   & 0.8                  & 0.6                     & 0.2                  & 0.4                   \\  
                        &                                   & 100\%                       & 0.4                  & 0.3                  & 0.8                  & 0.5                   & 0.7                  & 0.5                     & 0.2                  & 0.4                   \\ \hline
\end{tabular}
\end{table}

\clearpage

\subsection*{Node Visualization}
\begin{figure}[h]
    \centering
    \includegraphics[scale = 0.5]{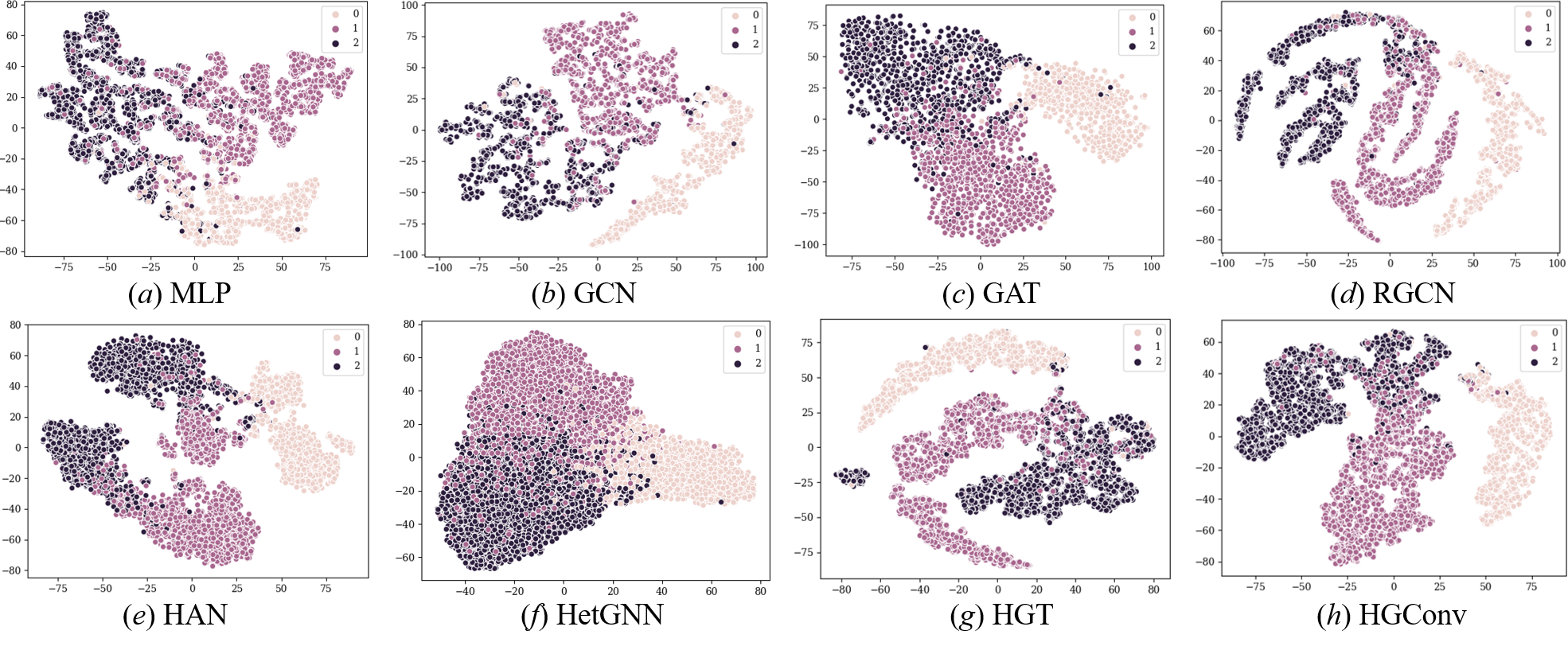}    
    \caption{Visualization of node representation on ACM-3.}
    \label{fig:acm_3_node_visualization}
\end{figure}

\begin{figure}[!ht]
    \centering
    \includegraphics[scale = 0.5]{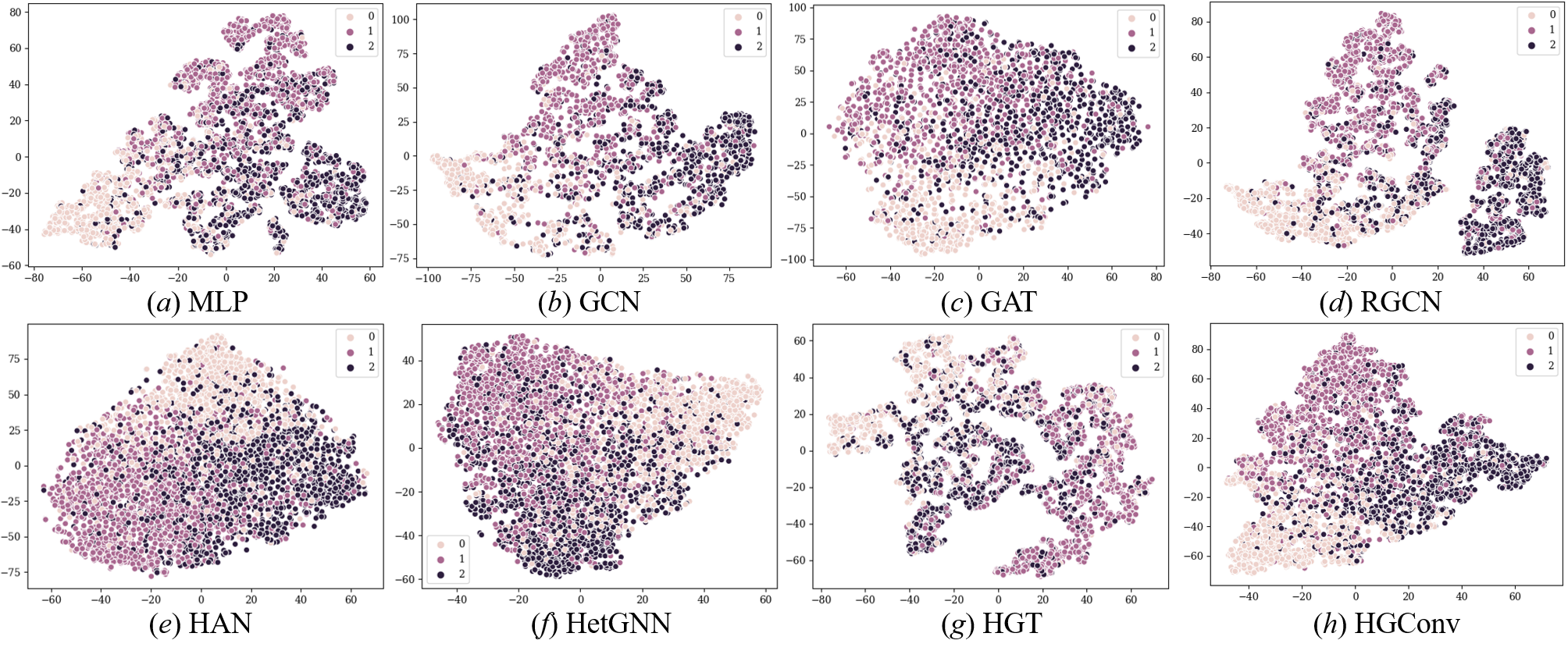}    
    \caption{Visualization of node representation on IMDB.}
    \label{fig:imdb_node_visualization}
\end{figure}

To show the effectiveness of the node representations, we conduct experiments on ACM-3 and IMDB datasets either. In \figref{fig:acm_3_node_visualization}, each point indicates a paper and its color denotes the published area. In \figref{fig:imdb_node_visualization}, each point indicates a movie and its color corresponds to the movie class. It can be known that our method could well separate the different categories of nodes. 

\vspace{0.5cm}

\vspace{1cm}

\end{document}